\documentclass[11pt]{article}
\usepackage{xspace, amsthm, amssymb, amsmath, enumitem}

\newcommand{\framework}{\textsc{Muon$^2$}\xspace}
\newcommand{\meframework}{\textsc{Muon$^2$-F}\xspace}
\newcommand{\muon}{\textsc{Muon}\xspace}
\newcommand{\mat}[1]{\mathbf{#1}}
\newcommand{\green}[1]{\textcolor{green!50!black}{#1}}

\newcommand{\down}[2]{\makecell{#1\\\red{(+#2)}}}
\newcommand{\up}[2]{{\bf#1} \green{(-#2)}}

\newtheorem{proposition}{Proposition}
\newtheorem{corollary}{Corollary}

\usepackage{algorithm}
\usepackage{algpseudocode}
\usepackage{graphicx}
\usepackage{subcaption}
\usepackage{booktabs} 
\usepackage{multirow}
\usepackage{makecell}
\usepackage[table]{xcolor}
\usepackage{array}
\usepackage{hyperref}
\usepackage[capitalize,noabbrev]{cleveref}
\usepackage{footnote}
\newcolumntype{B}{>{\columncolor{blue!12}}c}
\newcolumntype{b}{>{\columncolor{blue!6}}c}

\usepackage[final]{acl}

\usepackage{times}
\usepackage{latexsym}

\usepackage[T1]{fontenc}

\usepackage[utf8]{inputenc}

\usepackage{microtype}

\usepackage{inconsolata}

\usepackage{graphicx}

\title{\framework: Boosting \muon via Adaptive Second-Moment Preconditioning}

\author{
 \textbf{Ziyue Liu}\textsuperscript{1}, 
 \textbf{Ruijie Zhang}\textsuperscript{1}, 
 \textbf{Zhengyang Wang}\textsuperscript{1}, 
 \textbf{Yequan Zhao}\textsuperscript{1},
 \textbf{Yupeng Su}\textsuperscript{1},\\
 \textbf{Zi Yang}\textsuperscript{2},
 \textbf{Zheng Zhang}\textsuperscript{1}
\\
\textsuperscript{1}University of California at Santa Barbara;
\textsuperscript{2}University at Albany, SUNY \\
\{ziyueliu, zzhang01\}@ucsb.edu \\
\url{https://github.com/alvin-zyl/Muon2}
}

\begin{document}
\maketitle

\begin{abstract}
\muon has emerged as a promising optimizer for large-scale foundation model pre-training by exploiting the matrix structure of neural network updates through iterative orthogonalization. However, the orthogonalization quality of \muon hinges on the number of Newton--Schulz (NS) iterations performed, which poses efficiency challenges due to its non-trivial computation and communication cost. We propose \framework, an extension of \muon, to improve both quality and efficiency by applying Adam-style adaptive second-moment preconditioning before orthogonalization. Our key insight is that the core challenge of polar approximation in \muon lies in the ill-conditioned momentum matrix, of which the spectrum is substantially improved by \framework, leading to faster convergence toward a practically sufficient orthogonalization. We further characterize the practical orthogonalization quality via directional alignment, under which \framework demonstrates dramatic improvement over \muon at each polar step. Across GPT, LLaMA, and Mixture-of-Experts pre-training experiments up to 13B parameters, \framework (and its memory-efficient variant \meframework that preserves most of its benefits) consistently outperforms \muon and its variants while reducing NS iterations by 40\%, and saves up to 1/4 training time over \muon when achieving the same loss.
\end{abstract}

\section{Introduction}
The rapid progress of modern large-scale neural networks has been driven by the continual expansion of model capacity and training data~\cite{hoffmann2022training, kaplan2020scaling}. This paradigm has enabled the emergence of highly capable foundation models across language, vision, and multi-modal domains~\cite{achiam2023gpt, grattafiori2024llama, team2023gemini}. However, the increasing scale of these systems has made pre-training extremely resource-intensive, requiring vast computational budgets and long training durations. As a result, numerous efforts have been devoted to improve the efficiency of the pre-training system, spanning across model architectures~\cite{adler2024nemotron,liu-etal-2025-cola,zhang2025lax}, infrastructures~\cite{shoeybi2019megatron,rajbhandari2020zero,wang2025boost}, and optimization algorithms~\cite{gupta2018shampoo,vyas2024soap,jordan2024muon}.

Among existing optimization methods, adaptive first-order optimizers such as Adam~\cite{kingma2014adam} and AdamW~\cite{loshchilov2017decoupled} remain the de facto choice for training large models due to their robustness and ease of use. Nevertheless, their overlook of the underlying matrix structure of neural network parameters, have motivated substantial research into alternative optimization strategies~\cite{kfac,gupta2018shampoo,vyas2024soap}.

\muon~\cite{jordan2024muon} has emerged as a breakthrough that 
explicitly exploits the matrix structure of neural network gradients, without the full cost of computing second-order statistics. \muon approximates a polar decomposition of the momentum via the Newton--Schulz iteration to efficiently orthogonalize the update to mitigate gradient rank collapse and improve optimization dynamics in large models. Various studies have shown improved stability and overall performance by deploying \muon in large-scale foundation model pre-training~\cite{liu2025muonscalable,shah2025practical,zeng2025glm,team2025kimi}. 

Building on these successes, \muon remains an active area of research, with a number of recent works proposing variants that improve different aspects of the optimizer~\cite{khaled2025muonbp,li2025normuon,si2025adamuon,amsel2026polar,boissin2025turbo,ahn2025dion,zhang2026teon}. Most of these approaches explore modifications to \muon’s update rules, yet few has devoted to tackle the core challenge: a non-trivial amount of Newton–Schulz (NS) iterations per update that determines the orthogonalization quality and introduces computation and communication overhead. This raises a natural question: can we simultaneously enhance the quality and reduce the burden of \muon's orthogonalization procedure?

\vspace{-3pt}
\paragraph{Contributions} In this work, we investigate this question and introduce \framework, a simple yet effective modification of \muon that leverages adaptive second-moment scaling as an effective preconditioner for \muon's orthogonalization step. Our key observation is that applying Adam-style per-parameter scaling prior to the orthogonalization significantly improves the spectral properties of the momentum matrix. Empirically, this produces a more favorable singular value distribution that simultaneously improves the convergence of Newton–Schulz iterations and the final model performance. These improvements lead to an optimizer that is both computationally lighter and empirically stronger. We summarize our contributions:
\begin{itemize}[leftmargin=*]
    \item We propose \framework, a novel generalization of \muon optimizer that preconditions the momentum matrix via Adam-style adaptive scaling prior to \muon's orthogonalization. This simple yet effective approach simultaneously boosts model performance and training efficiency.
    \vspace{-5pt}
    \item To justify \framework, we identify the challenge of polar approximation lies in its input matrix's ill-conditioned spectrum and demonstrate that \framework significantly improves the input matrix conditioning, achieving superior directional alignment to the true orthogonalized update and substantially reducing polar iterations.
    \vspace{-5pt}
    \item We also propose \meframework, a memory-efficient version of \framework with a factorized second-moment preconditioner. This variant effectively reduces the memory overhead of saving the full second-moment while preserving most of \framework's performance gain.
    \vspace{-5pt}
    \item We conduct comprehensive experiments on pre-training GPT, LLaMA and Mixture-of-Experts models for up to 13B parameters. Experiments show that \framework family consistently outperforms \muon and its variants with 40\% fewer Newton--Schulz iterations, and reduce 1/4 training time for achieving the same loss.
\end{itemize}

\section{Related Work}

{\bf Coordinate-wise Adaptive Methods.} Despite ignoring underlying matrix structures, adaptive first-order optimizers remain the dominant choice for large-scale training. Adagrad~\cite{adagrad} introduced per-parameter adaptive scaling based on historical gradients, while Adam(W)~\cite{kingma2014adam,loshchilov2017decoupled} further incorporate exponential moving averages of first and second moments. To reduce memory overhead, Adafactor~\cite{shazeer2018adafactor} factorizes second-moment statistics, while more recent variants such as Adam-mini~\cite{adam-mini} and GaLore~\cite{zhao2024galore} simplify or compress optimizer states while retaining performance.

\vspace{-3pt}
\paragraph{\bf Matrix-Structured Methods.} An alternative line of work leverages matrix structure for improved conditioning. Shampoo~\cite{gupta2018shampoo} applies Kronecker-factored second-order preconditioning, while SOAP~\cite{vyas2024soap} stabilizes it with adaptive scaling. \muon~\cite{li2025normuon} operates directly on matrix-valued momentum, approximating its polar factor via iterative orthogonalization.

\vspace{-3pt}
\paragraph{\bf Variants of \muon.} Recent works extend \muon along multiple directions. PolarExpress~\cite{amsel2026polar} and Turbo-Muon~\cite{boissin2025turbo} improve the convergence of polar approximation, while NorMuon~\cite{li2025normuon} and AdaMuon~\cite{si2025adamuon} incorporate second-moment statistics into the update rule. Dion~\cite{ahn2025dion} explores low-rank orthogonalization for scalability in distributed settings. These methods improve either the efficiency or effectiveness of \muon, rather than jointly addressing both.

\section{The \framework\ Optimizer}

\subsection{Introducing \framework}
\begin{algorithm}[t]
\caption{The \framework Optimizer}
\label{alg:muon^2}
\begin{algorithmic}[1]
\Require 2D weights $\mathbf{W}_t \in \mathbb{R}^{n\times m}$, objective $\mathcal{L}$, learning rate $\eta$, momentum coefficients $\beta_1,\beta_2$, Newton--Schulz steps $K$, numerical constant $\epsilon$
\Ensure Updated weights $\mathbf{W}_{t+1}$
\State $\mathbf{M}_0 \gets \mathbf{0}$, $\mathbf{V}_0 \gets \mathbf{0}$
\For{$t \gets 1,2,\ldots$}
    \State $\mathbf{G}_t \gets \nabla_{\mathbf{W}_t}\mathcal{L}(\mathbf{W}_t)$
    \State $\mathbf{M}_t \gets \beta_1 \mathbf{M}_{t-1} + (1-\beta_1)\mathbf{G}_t$
    \State $\mathbf{V}_t \gets \beta_2 \mathbf{V}_{t-1} + (1-\beta_2)(\mathbf{G}_t \odot \mathbf{G}_t)$
    \State $\widetilde{\mathbf{M}}_t \gets \mathbf{M}_t \oslash (\sqrt{\mathbf{V}_t} + \epsilon \mathbf{1})$
    \State $\mathbf{O}_t \gets \mathrm{Newton\text{--}Schulz}(\widetilde{\mathbf{M}}_t, K)$
    \State $\mathbf{W}_{t+1} \gets \mathbf{W}_t - \eta\sqrt{m/n} \mathbf{O}_t$
\EndFor
\end{algorithmic}
\end{algorithm}

We introduce \framework, a novel generalization of \muon that integrates second-moment preconditioning prior to the orthogonalization step. The algorithm is summarized in \cref{alg:muon^2}.

Given a parameter matrix $\mathbf{W}_t \in \mathbb{R}^{n \times m}$ and gradient
\begin{equation}
\mathbf{G}_t = \nabla_{\mathbf{W}_t} \mathcal{L}(\mathbf{W}_t),
\end{equation}
\muon\ first constructs a momentum estimate
\begin{equation}
\mathbf{M}_t = \beta_1 \mathbf{M}_{t-1} + (1-\beta_1)\mathbf{G}_t.
\label{eq:momentum-ori}
\end{equation}

\framework\ augments this step with a second-moment accumulator
\begin{equation}
\mathbf{V}_t = \beta_2 \mathbf{V}_{t-1} + (1-\beta_2)(\mathbf{G}_t \odot \mathbf{G}_t),
\label{eq:second-moment}
\end{equation}
which produces a preconditioned momentum
\begin{equation}
\widetilde{\mathbf{M}}_t
=
\mathbf{M}_t \oslash (\sqrt{\mathbf{V}_t} + \epsilon\mathbf{1}),
\label{eq:momentum-ours}
\end{equation}
where $\odot$ and $\oslash$ denote element-wise multiplication and division, respectively. 

The preconditioned matrix is then orthogonalized using $K$ steps of the Newton--Schulz (NS) iteration
\begin{equation}
\mathbf{O}_t = \mathrm{NewtonSchulz}(\widetilde{\mathbf{M}}_t, K).
\end{equation}

Finally the parameter update is
\begin{equation}
\mathbf{W}_{t+1}
=
\mathbf{W}_t
-
\eta
\sqrt{\frac{m}{n}}
\,\mathbf{O}_t.
\end{equation}
where the $\sqrt{\frac{m}{n}}$ learning rate factor is proposed by~\cite{bernstein2025deriving} for better scalability.

Compared with \muon, the only modification introduced by \framework\ is the second-moment scaling prior to orthogonalization, as shown in Eq.~\eqref{eq:momentum-ours}. As we will show in the following sections, this simple yet effective modification substantially improves the spectral properties of the matrix entering the NS iteration, enabling better and faster convergence of the orthogonalization procedure.

\subsection{Revisiting Polar Approximation in \muon}

A central component of \muon\ is the use of the Newton--Schulz iteration to approximate the polar factor of a matrix.\footnote{Strictly speaking, this map is an \emph{orthonormalization}; we follow common usage and write ``orthogonalization''.} Existing theoretical analyses of polar methods~\cite{amsel2026polar,chen2014stable,grishina2025accelerating} typically measure approximation quality through \emph{exact orthogonality}, for example via quantities such as
\begin{equation}
\|\mat{Q}^\top \mathbf{Q}- \mathbf{I}\|,
\label{eq:exact_orth}
\end{equation}
which quantifies how close the approximate factor $\mat{Q}$ is to an orthogonal matrix\footnote{We assume matrices have no fewer rows than columns; \muon's implementation transposes wide matrices before orthogonalization anyway.}.

However, we argue that despite of its unwavering mathematical correctness, this notion of quality does not fully characterize the role polar approximation plays in \muon-family optimizers. In particular, the original \muon work~\cite{jordan2024muon} explicitly uses an \emph{inexact} orthogonalization that roughly maps singular values to $[1-\epsilon, 1+\epsilon]$. And surprisingly, $\epsilon$ can be as large as $\sim 0.3$ without harming the performance of \muon. 



Let's pivot to another example showcasing why Eq.~\eqref{eq:exact_orth} may fail as a practically effective metric. Consider a scenario where all singular values are projected to an exact constant $c \in (0,1]$. Under this construction, the resulting matrix can be written as
\begin{equation}
\mat{Q} = c\,\mathbf{U}\mathbf{V}^\top,
\label{eq:scaled-polar}
\end{equation}
where $\mathbf{U}\mathbf{V}^\top$ is the exact polar factor from singular value decomposition (SVD) 
\begin{equation}
    \mat{G} = \mat{U\Sigma V^T}, \ \mat{Q}_\star = \mat{UV^T}.
\end{equation}
And Eq.~\eqref{eq:exact_orth} becomes
\begin{equation}
\|\mat{Q}^\top \mat{Q} - \mathbf{I}\|
= |c^2 - 1| \cdot \|\mathbf{I}\|.
\end{equation}
Therefore, the orthogonality error depends entirely on the deviation of $c$ from $1$. In particular, even if the matrix preserves the exact singular directions, any global scaling $c \neq 1$ leads to a large orthogonality error despite $\mat{Q}$ being perfectly aligned with the true polar factor.

We remark that for optimization, the approximate orthogonalized matrix is not used as a standalone object but rather as the \emph{update direction}, where
\begin{equation}
\Delta \mathbf{W} = -\eta \mat{Q},
\end{equation}
and any scaling factor $c$ that $\mat{Q}$ may possess will be absorbed into the step size $\eta$ and being tuned as a hyper-parameter in practice. Therefore, a scale dependent metric such as Eq.~\eqref{eq:exact_orth} does not fully capture what orthogonalization achieves in practice and could be misleading in certain cases. 

\subsection{Directional Alignment}
In contrast, if we were to measure directional alignment instead of exact orthogonality, cosine similarity becomes a strong candidate as it cancels the scaling effect on each singular value, i.e.,
\begin{equation}
\frac{\langle \mat{Q}, \mathbf{Q}_\star \rangle_F}
{\|\mat{Q}\|_F \|\mathbf{Q}_\star\|_F}
=
\frac{c \|\mathbf{Q}_\star\|_F^2}
{|c| \|\mathbf{Q}_\star\|_F^2}
= 1.
\end{equation}
Thus, cosine similarity\footnote{The matrix inner product is $\langle \mat{A}, \mathbf{B}\rangle_F = \mathrm{Tr}(\mat{A}^T\mat{B})$.} is invariant to the global scaling factor $c$ and reflects the fact that the update direction is unchanged. 

More generally, within the Newton--Schulz (NS) iteration that \muon~\cite{jordan2024muon} applies, running one step yields
\begin{equation}
\begin{aligned}
    \mat{Q}_{\text{NS}}^{(1)} &= a\mat{G} + b (\mat{GG}^T)\mat{G} + c(\mat{GG}^T)^2\mat{G} \\
    &= \mat{U}(a\mat{\Sigma} + b\mat{\Sigma}^3 + c\mat{\Sigma}^5)\mat{V}^T \\
    &= \mathbf{U}\,\mathrm{diag}(\phi(\sigma_1),\dots,\phi(\sigma_n))\,\mathbf{V}^\top
\end{aligned}
\label{eq:ns-iteration}
\end{equation}
where $\mat{G}=\mat{U\Sigma V}^T$ is the SVD of the momentum matrix, $\phi(x)=ax+bx^3+cx^5$ with coefficients $(a, b, c) = (3.4445, -4.7750, 2.0315)$ transforms each singular value $\sigma_i$. \muon repeats Eq.~\eqref{eq:ns-iteration} by five times, yielding
\begin{equation}
\mat{Q}_{\text{NS}}^{(5)} = \mathbf{U}\,\mathrm{diag}(\phi^5(\sigma_1),\dots,\phi^5(\sigma_n))\,\mathbf{V}^\top
\end{equation}
In this case, the cosine similarity between the NS output and the true polar factor becomes
\begin{equation}
\cos(\mathbf{Q}_{\text{NS}^{(5)}}, \mathbf{Q}_\star)
=
\frac{\sum_i \phi^5(\sigma_i)}
{\sqrt{n}\sqrt{\sum_i \phi^5(\sigma_i})^2},
\label{eq:cos-sim}
\end{equation}
which depends only on the relative singular value distribution of the NS output matrix and is invariant to scale-dependent statistics.

We remark that the cosine similarity we promote [Eq.~\eqref{eq:cos-sim}] does not necessarily contradict with the exact orthogonalization error [Eq.~\eqref{eq:exact_orth}] in practice, but cosine similarity does have superior robustness, interpretability and their practical applicability. We refer detailed discussions to \cref{apx:detail-cos-sim}.

\subsection{Spectral Effects of \framework}     

We now investigate why \framework can jointly improve the efficiency and effectiveness of \muon. We center around the Newton--Schultz (NS) iteration for our analysis as it is the only major complexity that \muon introduces over SGD/Adam-family optimizers. As shown by Eq.~\eqref{eq:ns-iteration}, each NS step introduces multiple matrix multiplications per parameter that require accessing the full-matrix on each device. This introduces not just computational overhead, but also communicational overhead that's often non-trivial and latency-bound in large-scale distributed setting. Therefore, it is essential to reduce the necessary NS steps for developing efficient \muon, yet the convergence of NS iteration largely depends on the steps being performed. To understand how \framework breaks free from this limitation, we analyze it in two-fold, focusing on how \framework changes: (1) The momentum matrix prior to the NS iteration. (2) The output of each NS iteration step.

For consistency, all quantitative studies in this section are conducted on training data collected from a LLaMA-60M model with \muon and \framework using the polar approximation defined in~\cite{jordan2024muon}. We argue that the claims and observations we make also generalize to other polar methods such as PolarExpress~\cite{amsel2026polar}, with minor differences in certain numerical values, see full details in \cref{apx:zone-polarexpree}.

\subsubsection{\framework Improves NS Input Spectrum}
\label{sec:input-spectrum}

\begin{figure}[]
    \centering
    \begin{subfigure}[t]{0.49\columnwidth}
        \centering
        \includegraphics[width=\linewidth]{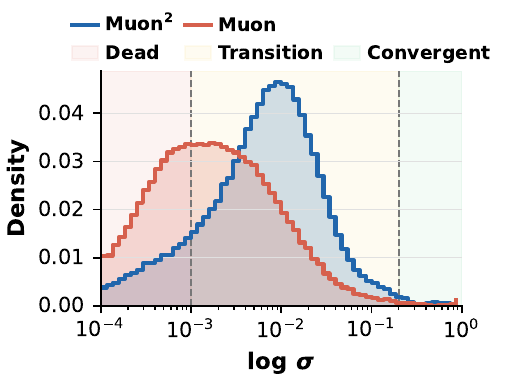}
        \caption{Early-training spectrum}
        \label{fig:ns-in-sv-dist-early}
    \end{subfigure}
    \hfill
    \begin{subfigure}[t]{0.49\columnwidth}
        \centering
        \includegraphics[width=\linewidth]{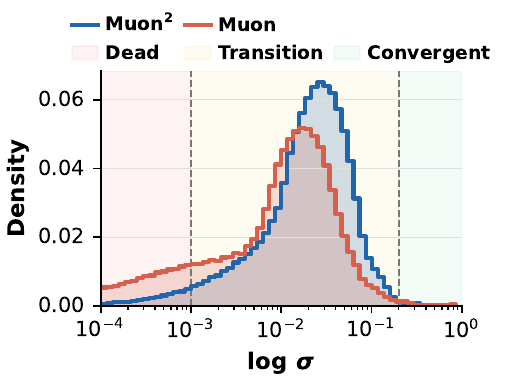}
        \caption{Mid-training spectrum}
        \label{fig:ns-in-sv-dist-mid}
    \end{subfigure}

    \vspace{0.5em}
    \begin{subfigure}[t]{0.49\columnwidth}
        \centering
        \includegraphics[width=\linewidth]{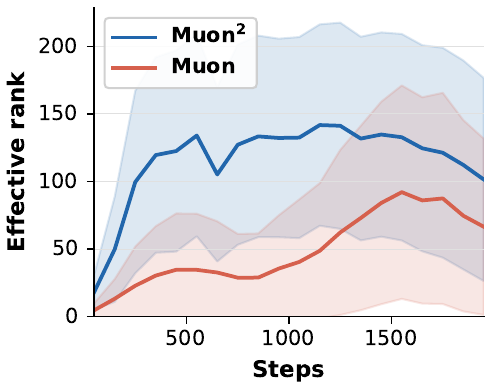}
        \caption{Effective rank}
        \label{fig:ns-in-eff-rank}
    \end{subfigure}
    \hfill
    \begin{subfigure}[t]{0.49\columnwidth}
        \centering
        \includegraphics[width=\linewidth]{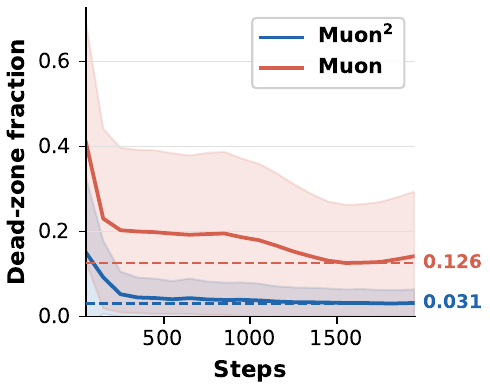}
        \caption{Dead zone fraction}
        \label{fig:dead-zone-frac}
    \end{subfigure}

    \caption{Spectral effect of \framework on the input matrix of the Newton--Schultz iteration.}
    \label{fig:ns-in-spec}
    \vspace{-10pt}
\end{figure}

\begin{figure}[t]
    \centering
    \begin{subfigure}[t]{0.49\columnwidth}
        \centering
        \includegraphics[width=\linewidth]{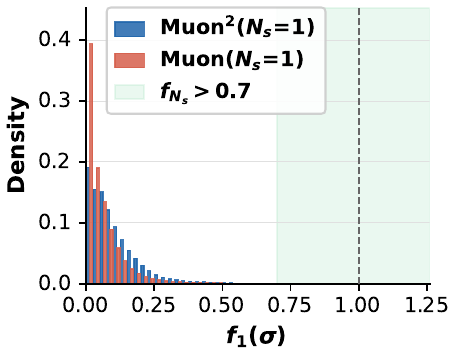}
        \caption{Spectrum at $N_s=1$}
        \label{fig:ns-1-out}
    \end{subfigure}
    \hfill
    \begin{subfigure}[t]{0.49\columnwidth}
        \centering
        \includegraphics[width=\linewidth]{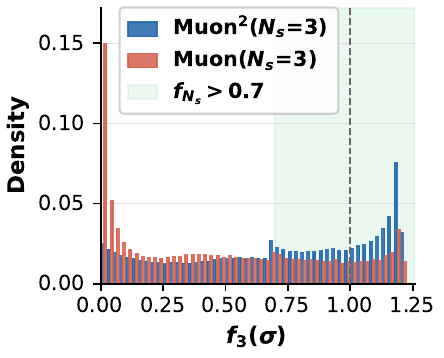}
        \caption{Spectrum at $N_s=3$}
        \label{fig:ns-3-out}
    \end{subfigure}

    \vspace{0.5em}
    \begin{subfigure}[t]{0.49\columnwidth}
        \centering
        \includegraphics[width=\linewidth]{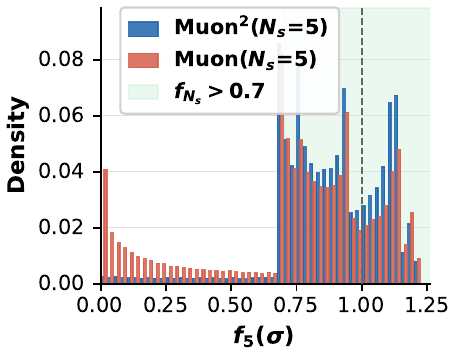}
        \caption{Spectrum at $N_s=5$}
        \label{fig:ns-5-out}
    \end{subfigure}
    \hfill
    \begin{subfigure}[t]{0.49\columnwidth}
        \centering
        \includegraphics[width=\linewidth]{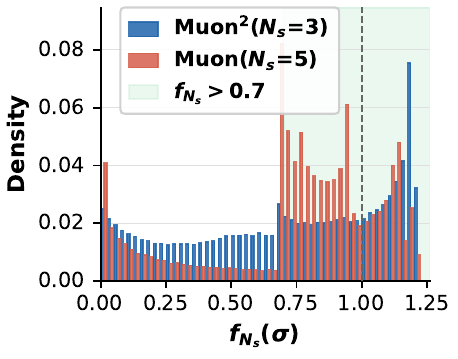}
        \caption{Spectrum at $N_s=3$ vs. 5}
        \label{fig:ns-3-5-out}
    \end{subfigure}

    \caption{Spectral effect of \framework on the Newton--Schultz (NS) output (i.e., ${\bf f_{N_s}}(\sigma)$) at each step (i.e., $N_s$).}
    \label{fig:ns-out-sv-dist}
    \vspace{-10pt}
\end{figure}

We start with characterizing the singular value distribution of the input momentum matrix of NS, that is Eq.~\eqref{eq:momentum-ori} for \muon and Eq.~\eqref{eq:momentum-ours} for \framework. We normalize it by its Frobenius norm~\cite{jordan2024muon} to reflect the actual input of the NS iteration. Given an $\epsilon$, which defines the practical orthogonalization target of mapping any normalized singular values $\sigma_i$ to $[1-\epsilon, 1+\epsilon]$, and given a practical choice of Newton-Schultz steps $N_s$, we define the following convergence zones for a polar approximation based on its destination values $\phi^{N_s}_{\epsilon}(\sigma_i)$:
\begin{itemize}[leftmargin=*]
\vspace{-5pt}
    \item \emph{Dead zone}: $\phi^{N_s}_{\epsilon}(\sigma_i)<1-\epsilon$, reflects the range of singular values that fail to converge after $N_s$ steps. Larger the dead zone is, more deviated the results are from the true orthogonalized update. 
\vspace{-5pt}
    \item \emph{Transition zone}: $\phi^{N_s}_{\epsilon}(\sigma_i)\geq 1-\epsilon$ and  $\phi^{1}_{\epsilon}(\sigma_i)<1-\epsilon$. This is the region where non-trivial amount of $N_s$ steps are needed to achieve the practical orthogonalization target.
\vspace{-5pt}
    \item \emph{Convergent zone}: $\phi^{1}_{\epsilon}(\sigma_i)\geq 1-\epsilon$, where $\sigma_i$ is large enough that only one NS iteration is needed.
\end{itemize}
Concretely, under \muon's setting where $\epsilon\approx0.3$, $N_s=5$, we can calculate each zone roughly as $[0, 0.001]$, $[0.001, 0.2]$, $[0.2, 1]$, and is colored differently in \cref{fig:ns-in-sv-dist-early,fig:ns-in-sv-dist-mid}.

The biggest practical challenge for NS is to project a wide range of singular values that spans multiple orders of magnitude to close one as fast as possible. As shown in \cref{fig:ns-in-sv-dist-early}, \muon's early-training stage singular value distribution for NS input spans from $10^{-4}$ to 1, and is centered around $10^{-3}$, where almost half singular values fall into the dead zone. Comparatively, \framework shows a significantly right-shifted distribution, where it centers around $10^{-2}$, $10\times$ larger than \muon, and the majority falls into the transition zone. As the training continues, both methods have their distribution shifting right, while \framework shows consistently lower fraction in the dead zone, demonstrated by \cref{fig:ns-in-sv-dist-mid,fig:dead-zone-frac}. In addition to lower dead-zone fraction, \cref{fig:ns-in-eff-rank} shows that \framework has consistently higher effective rank throughout training, highlighting the fact that its singular values are closer together, i.e., a tighter spectrum, which coincides with the properties that cosine similarity [Eq.~\eqref{eq:cos-sim}] promotes. Therefore, we anticipate the spectrum after polar transformation will also be tighter, resulting a higher cosine similarity, thus more aligned with the true orthogonalized update.

These findings conclude our first perspective: the preconditioning effect of \framework significantly improves the input spectrum of the NS iteration, providing a stronger baseline that is easier to achieve practically sufficient polar approximation.

\subsubsection{\framework Improves Polar Quality}

Now we compare the polar approximation behavior of \framework and \muon by visualizing their singular value distributions at NS steps ($N_s$) 1, 3, and 5 in \cref{fig:ns-1-out,fig:ns-3-out,fig:ns-5-out}. Across all cases, \framework exhibits a tighter spectrum with substantially fewer near-zero singular values. By $N_s=5$, its singular values fall almost entirely within the target range. We further compare \framework at $N_s=3$ against \muon at $N_s=5$ in \cref{fig:ns-3-5-out}. Despite using substantially fewer iterations, \framework produces a highly similar distribution while reducing the density of extremely small singular values by roughly half. These findings suggest that \framework at $N_s=3$ already provides a strong polar approximation.

\begin{figure}[t]
    \centering
    \includegraphics[width=\linewidth]{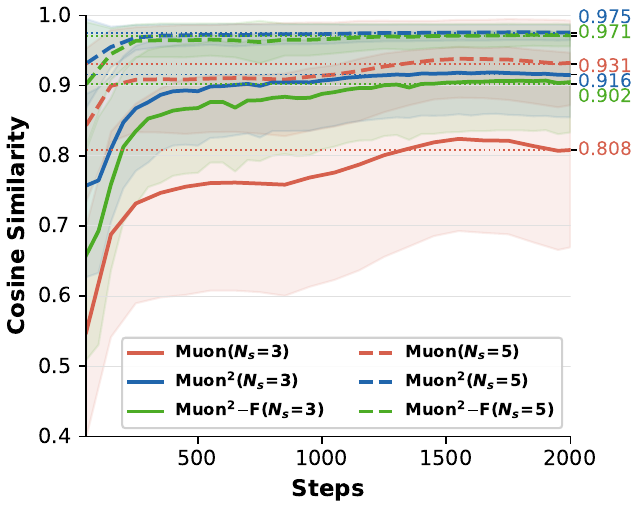}
    \caption{Cosine similarity of the output matrix vs. true orthogonalized update for \muon, \framework and \meframework at different Newton-Schultz steps.}
    \label{fig:cos-sim-trend}
    \vspace{-10pt}
\end{figure}

Quantitatively, we calculate the cosine similarity [Eq.~\eqref{eq:cos-sim}] of \framework and \muon at each $N_s$, reflecting how closely the approximate polar factors align with the true orthogonalized gradients. As shown in \cref{fig:cos-sim-trend}, \framework at $N_s=3$ already achieves cosine similarity comparable to \muon at $N_s=5$ (0.916 vs.\ 0.931), and further improves to 0.975 at $N_s=5$, substantially surpassing \muon. In contrast, reducing \muon from $N_s=5$ to $N_s$ $=3$ causes a significant drop in cosine similarity (0.931 $\rightarrow$ 0.808), which we later show also leads to a dramatic degradation in model performance.

To better connect the qualitative and quantitative analyses, we compare \framework at $N_s=3$ with \muon at $N_s=5$ in \cref{fig:ns-3-5-out}. The visualization shows that \muon has monotonically decreasing densities between near-zero and the target range, whereas \framework exhibits a more uniform spread with higher density in the transition and convergent zones. This suggests a more practically favored spectrum of \framework that accounts for its further improvement at $N_s=5$.

These findings conclude our second perspective: benefited from the spectral effect of the second moment preconditioning, \framework consistently produces updates that are both better aligned with and easier to progress towards the true orthogonalized gradient at every NS iteration. This effectively reduces the necessary steps of the NS iteration and substantially improves the polar approximation quality.

\subsubsection{A Random-Matrix Perspective}
\label{sec:theory}
The two perspectives above are established empirically. In this section, we further provide a theoretical account for the spectral effect of \framework. We model the momentum matrix as $M_{ij} = s_{ij}\,\xi_{ij}$, where the deterministic scale $s_{ij} > 0$ captures the persistent fluctuation magnitude of each entry, and $\xi_{ij}$ are independent zero-mean, unit-variance variables. Under this model, the spectrum of $M$ is determined by the scale profile $(s_{ij})$: when the scales are highly uneven, a large fraction of singular values can be pushed towards zero (we give a concrete construction in \cref{apx:proofs}), which is exactly the ill-conditioned NS input that \cref{sec:input-spectrum} identifies as the practical challenge. Meanwhile, the preconditioning of \framework [Eq.~\eqref{eq:momentum-ours}] divides each entry by the second-moment estimate of precisely this scale. The following proposition characterizes the spectrum after the division.

\begin{proposition}
\label{prop:mp}
Let $(\xi_{ij})_{i,j \ge 1}$ be a fixed infinite array of i.i.d.\ variables with zero mean, unit variance, and $\mathbb{E}\xi^4 < \infty$. For a sequence $m = m_n \le n$ with $m/n \to \lambda \in (0,1)$, let $M = (s_{ij}\,\xi_{ij}) \in \mathbb{R}^{m\times n}$, where the scales $s_{ij} > 0$ are deterministic, and let $\widetilde{M} := (M_{ij}/s_{ij})$ denote the exactly preconditioned matrix. Then the empirical singular-value distribution of $\widetilde{M}/\sqrt{n}$ converges almost surely to the singular-value form of the Marchenko--Pastur law with ratio $\lambda$~\citep{marvcenko1967distribution} --- the pushforward of the Marchenko--Pastur law under $x \mapsto \sqrt{x}$, supported on $[1-\sqrt{\lambda},\, 1+\sqrt{\lambda}]$ --- independently of $(s_{ij})$. Moreover, the empirical distribution of $\{\sqrt{m}\,\sigma_i(\widetilde{M}/\|\widetilde{M}\|_F)\}_{i=1}^m$ converges almost surely to the same law; consequently, for every $0 \le t < 1-\sqrt{\lambda}$, the fraction of singular values of the Frobenius-normalized NS input below $t/\sqrt{m}$ vanishes almost surely.
\end{proposition}

\cref{prop:mp} states that once the scales are divided out, the singular values concentrate on the Marchenko--Pastur support $[1-\sqrt{\lambda}, 1+\sqrt{\lambda}]$, independently of $(s_{ij})$. In other words, the preconditioning removes the dependence of the spectrum on the scale profile: no matter how uneven the scales are, the fraction of singular values below the lower edge $1-\sqrt{\lambda}$ vanishes. This edge depends only on the aspect ratio of the layer: it is positive for rectangular layers such as the MLP projections, which is why the proposition assumes $\lambda < 1$, while for square layers it sits at zero and a weaker form of the same conclusion holds (see \cref{apx:proofs}). Consequently, the dead-zone fraction of the preconditioned input vanishes, matching the right-shifted, tighter spectrum observed in \cref{sec:input-spectrum}.

In practice, the momentum is not pure noise but also accumulates a signal component, which corresponds to a nonzero mean $\mu_{ij}$ in our model. The following corollary shows that the conclusion of \cref{prop:mp} is robust to such a signal in two regimes: a low-rank signal of any bounded strength, and a weak signal of any rank.

\begin{corollary}
\label{cor:signal}
In the setting of \cref{prop:mp}, let $M_{ij} = \mu_{ij} + s_{ij}\xi_{ij}$ with deterministic $\mu_{ij}$, so that $\widetilde{M} = D + \Xi$ with $D := (\mu_{ij}/s_{ij})$ and $\Xi := (\xi_{ij})$. (i) If $\operatorname{rank}(D) = o(m)$ and $\limsup_n \|D\|_F^2/(mn) \le d^2 < \infty$ for some $d \ge 0$, then almost surely the fraction of NS-input singular values below $t/\sqrt{m}$ vanishes for every $0 \le t < (1-\sqrt{\lambda})/(1+d)$. (ii) If $\|D\|_F^2 = o(mn)$, then all conclusions of \cref{prop:mp} hold unchanged.
\end{corollary}

In the corollary, $D$ is the matrix of per-entry signal-to-noise ratios. Case (i) corresponds to the regime suggested by our measurements: the low effective ranks in \cref{sec:input-spectrum} indicate a signal concentrated in few directions, and for such a signal the guaranteed margin only degrades from $1-\sqrt{\lambda}$ to $(1-\sqrt{\lambda})/(1+d)$. Case (ii) shows that a signal spread over many directions is tolerated as long as it is weak. We note the limitations of these results: they are asymptotic, they control the fraction of singular values in the dead zone rather than the smallest one, and they idealize the second-moment estimate as the exact scale. We quantify these gaps, together with the connection to the factorized second moment of \meframework, in \cref{apx:proofs}, where the proofs are also given.

\subsection{Overall Benefits of \framework}

{\bf \framework delivers a stronger orthogonalization.} With the same number of Newton-Schultz (NS) iterations, \framework significantly improves the alignment of \muon's update direction with the true orthogonalized gradient.

\vspace{1pt}
\noindent {\bf \framework reduces NS iterations for practically sufficient orthogonalization.} To achieve a similar level of update-direction alignment, \framework reduces the required NS iterations by 40\%. We will further show in the experiments that 3-step \framework already achieves better model performance than 5-step \muon. We speculate that this is due to the richer historical information carried by our second-moment preconditioning.

\subsection{\meframework: A Memory Efficient \framework}



From Eq.~\eqref{eq:second-moment} and \eqref{eq:momentum-ours}, \framework introduces additional memory overhead by maintaining a second moment throughout training. However, since the second moment only conditions the input to orthogonalization rather than directly determining the update, we expect \framework to be relatively insensitive to approximation error. Motivated by this, we adopt Adafactor-style factorized second moments:
\begin{equation}
\begin{aligned}
    \mathbf{r}_t &\gets \beta_2 \mathbf{r}_{t-1} + (1-\beta_2)\,\mathrm{sum}_{\mathrm{col}}(\mathbf{R}_t),\\
    \mathbf{c}_t &\gets \beta_2 \mathbf{c}_{t-1} + (1-\beta_2)\,\mathrm{sum}_{\mathrm{row}}(\mathbf{R}_t), \\
    \widehat{\mathbf{V}}_t &\gets \dfrac{\mathbf{r}_t\,\mathbf{c}_t^\top}{\mathrm{sum}(\mathbf{r}_t)}.
\end{aligned}
\label{eq:adafactor}
\end{equation}
Instead of storing the full second-moment matrix, this approximation keeps only row- and column-wise statistics and reconstructs the matrix via their outer product. We denote this variant as \framework-Factorized (\meframework). As shown later, \meframework remains close to the exact \framework in both reduced NS iteration requirements and model performance, while keeping memory overhead nearly unchanged from \muon. Full details are in \cref{alg:muon2_adafactor}.

\section{Experiments}
\begin{table}[t]
\centering
\footnotesize
\setlength{\tabcolsep}{5pt}
\resizebox{\linewidth}{!}{%
\begin{tabular}{c c c B b}
\toprule
\textbf{Model} & {\bf NS Steps} & {\bf \muon} & \cellcolor{blue!12}{\bf \framework} & {\bf \meframework} \\
\midrule

\multirow{2}{*}{GPT-Small}
& $N_s=3$ & 32.70 & \up{28.12}{4.58} & \up{28.30}{4.40}  \\
& $N_s=5$ & 29.51 & \up{27.95}{1.56} & \up{27.93}{1.58}\\
\midrule

\multirow{2}{*}{GPT-Base}
& $N_s=3$ & 24.69 & \textbf{20.39} \green{(-4.30)} & \up{21.17}{3.52} \\
& $N_s=5$ & 21.47 & \textbf{19.96} \green{(-1.51)} & \up{20.58}{0.89} \\
\midrule

\multirow{2}{*}{GPT-Large}
& $N_s=3$ & 21.13 & \textbf{16.99} \green{(-4.14)} & \up{17.69}{3.44} \\
& $N_s=5$ & 17.56 & \textbf{16.52} \green{(-1.04)} & \up{16.55}{1.01} \\
\bottomrule
\end{tabular}
}
\caption{Validation perplexity ($\downarrow$) of \muon and \framework on GPT models across Newton--Schulz iterations.}
\label{tab:gpt_main_results}
\vspace{-10pt}
\end{table}

In this section, we evaluate \framework on extensive pre-training experiments that cover GPT, LLaMA and Mixture-of-Experts architectures on various scales. These experiments strongly and consistently demonstrate that \framework achieves two benefits: (1) Significantly better model performance; (2) Substantially fewer Newton--Schulz (NS) iterations. We also compare \framework against other \muon variants and show that none of them can achieve both benefits as \framework does.

\subsection{Pre-Training GPT}
\label{sec:pretrain-gpt}

We pre-train GPT models at three scales: small, base and large, with respectively 3.0B, 7.2B, and 15.5B tokens, following the compute-optimal training regime~\cite{hoffmann2022training}. We use the FineWeb dataset~\cite{penedo2024fineweb} tokenized by the GPT tokenizer. Trainings are run on H100/A100 GPUs, with the pipeline adapted from NanoGPT~\cite{nanogpt}. We use the polar method from the original \muon work~\cite{jordan2024muon}. Detailed configurations and hyper-parameters are provided in \cref{apx:gpt-sweep}.

\begin{figure}[t]
    \centering
    \includegraphics[width=\linewidth]{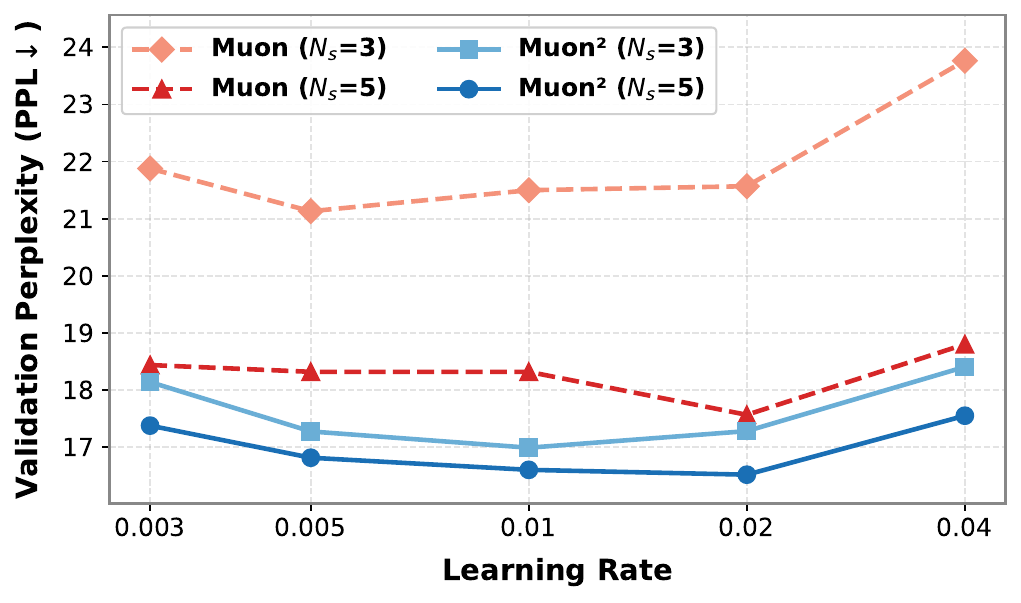}
    \caption{Learning rate sweep on GPT-Large comparing \framework and \textsc{Muon} across Newton--Schulz iterations.}
    \vspace{-5pt}
    \label{fig:gpt_large_sweep}
\end{figure}

As shown in \cref{tab:gpt_main_results}, \framework consistently outperforms \muon not just when they use the same NS steps, but also when \framework uses substantially fewer NS steps. With only $N_s=3$, \framework outperforms \muon with $N_s=5$ at all three scales. When both at $N_s=3$, the gap between \framework and \muon is even more dramatic. Meanwhile, the memory efficient version \meframework also achieves comparatively good performance with practically negligible memory cost. To ensure fair comparison, we sweep learning rates for \muon and \framework and show results at GPT-Large scale in \cref{fig:gpt_large_sweep}. We can observe that the benefits \framework provides are irrelevant to each individual choice of learning rate, suggesting its broad practical applicability.

\subsection{Pre-Training LLaMA}
\label{sec:pretrain-llama}
\begin{table}[t]
\centering
\footnotesize
\setlength{\tabcolsep}{5pt}
\resizebox{\linewidth}{!}{%
\begin{tabular}{c c c B b}
\toprule
\textbf{Model} & {\bf NS Steps} & {\bf \muon} & {\bf \framework} & {\bf \meframework} \\
\midrule

\multirow{2}{*}{LLaMA-60M}
& $N_s=3$ & 26.37 & \textbf{24.59} \green{(-1.78)} & {\bf 24.68} \green{(-1.69)} \\
& $N_s=5$ & 24.98 & \textbf{24.60} \green{(-0.38)} & {\bf 24.66} \green{(-0.32)} \\
\midrule

\multirow{2}{*}{LLaMA-350M}
& $N_s=3$ & 14.91 & \up{13.44}{1.47} & \up{13.55}{1.36} \\
& $N_s=5$ & 14.03 & \up{13.46}{0.57} & \up{13.44}{0.59} \\
\midrule

\multirow{2}{*}{LLaMA-1B}
& $N_s=3$ & 11.63 & \up{10.42}{1.21} & \up{10.49}{1.14} \\
& $N_s=5$ & 10.62 & \up{10.21}{0.41} & \up{10.21}{0.41} \\
\bottomrule
\end{tabular}
}
\caption{Validation perplexity ($\downarrow$) of \muon and \framework on LLaMA models across Newton--Schulz iterations.}
\label{tab:llama_main_results}
\vspace{-10pt}
\end{table}

We pre-train LLaMA-style models at four scales: 60M, 350M, 1B and 13B, with respectively 1.0B, 7.3B, 20.1B, and 26.8B. We use C4~\cite{c4} dataset tokenized by LLaMA-2 tokenizer. We adopt the training pipeline from Nanotron~\cite{nanotron}. Detailed configurations and hyper-parameters are provided in \cref{apx:llama-sweep}.

As shown in \cref{tab:llama_main_results,tab:13b-and-moe-results}, the results are consistent with GPT models. \framework continues to outperform \muon even with fewer NS iterations. At all four scales, reducing NS steps will cause a dramatic degradation for \muon but only minimum drops for \framework, and the gap between \framework and \meframework is also less pronounced.

\subsection{Pre-Training Mixture-of-Experts}
\label{sec:pretrain-moe}
\begin{table}[t]
\centering
\footnotesize
\setlength{\tabcolsep}{5pt}
\begin{tabular}{c c c B}
\toprule
\textbf{Model} & {\bf NS Steps} & {\bf \muon} & {\bf \framework} \\
\midrule

\multirow{2}{*}{LLaMA-13B}
& $N_s=3$ & 10.54 & \textbf{9.11} \green{(-1.43)} \\
& $N_s=5$ & 9.24 & \textbf{8.71} \green{(-0.53)} \\
\midrule

\multirow{2}{*}{MoE-7B-A1B}
& $N_s=3$ & 11.93  & \up{9.85}{2.08}  \\
& $N_s=5$ & 10.61  & \up{9.66}{0.95}  \\
\bottomrule
\end{tabular}
\caption{Validation perplexity ($\downarrow$) of \muon and \framework on LLaMA-13B and MoE models across NS iterations.}
\label{tab:13b-and-moe-results}
\vspace{-10pt}
\end{table}

We further evaluate on Mixture-of-Experts (MoE), a family of model architectures that is widely adopted in industry~\cite{achiam2023gpt,zeng2025glm,liu2024deepseek}. We pre-train a 7B-A1B MoE (1B active parameters, 2 out of total 16 experts) for 20.1B tokens on 4 variants: \muon and \framework each with $N_s=3$ and $N_s=5$ (\cref{tab:13b-and-moe-results}). \framework continues to outperform \muon significantly on both NS settings, and reduces NS steps by $40\%$ while improving model performance.

\subsection{Downstream Performance Evaluation}
\begin{table*}[t]
\centering
\small
\setlength{\tabcolsep}{6pt}
\renewcommand{\arraystretch}{1.15}
\resizebox{\linewidth}{!}{%
\begin{tabular}{lcccccccccc}
\toprule
& {\bf NS Steps} & \textbf{Avg} & \textbf{ARC-c}$^\dagger$ & \textbf{ARC-e} & \textbf{OBQA} & \textbf{HellaS.} & \textbf{PIQA} & \textbf{WinoG.}$^\dagger$ & \textbf{LAMBADA} & {\bf MMLU}$^\dagger$ \\

\midrule

\multirow{2}{*}{\muon} & $N_s=3$ & 42.56 & 27.08 & 42.80 & 32.00 & 44.37 & 71.22 &
56.75 & 36.52 & 29.75 \\
& $N_s=5$ & 45.36 & 30.01 & 46.09 & 33.20 & 49.75 & 72.25 & 59.91 & 40.99 & 30.71 \\
\midrule

\multirow{2}{*}{\framework} & \cellcolor{blue!6}$N_s=3$ & \cellcolor{blue!6}45.89 & \cellcolor{blue!6}31.91 & \cellcolor{blue!6}47.81 & \cellcolor{blue!6}31.80 & \cellcolor{blue!6}50.23 & \cellcolor{blue!6}72.20 & \cellcolor{blue!6}59.88 & \cellcolor{blue!6}42.50 & \cellcolor{blue!6}30.79 \\
& \cellcolor{blue!12}$N_s=5$ & \cellcolor{blue!12}\textbf{47.83} & \cellcolor{blue!12}\textbf{33.48} & \cellcolor{blue!12}\textbf{49.03} & \cellcolor{blue!12}\textbf{35.20} & \cellcolor{blue!12}\textbf{52.61} & \cellcolor{blue!12}\textbf{74.05} &
\cellcolor{blue!12}\textbf{61.90} & \cellcolor{blue!12}\textbf{45.08} & \cellcolor{blue!12}\textbf{31.31} \\
\bottomrule

\end{tabular}
}
\vspace{-5pt}
\caption{Downstream evaluation on LLaMA-1B across NS iterations. Tasks with $^\dagger$ are fine-tuned first (\cref{apx:downstream}); the rest are zero-shot.}
\label{tab:downstream-llama}
\vspace{-5pt}
\end{table*}

Beyond validation perplexity, we also verify the effectiveness of \framework by performing downstream evaluations on the trained LLaMA-1B in \cref{tab:downstream-llama}. Results are well aligned with validation perplexity that \framework consistently outperforms \muon. Full details and more evaluations are in \cref{apx:downstream}.

\subsection{Comparisons against \muon Variants}
We have demonstrated that \framework consistently outperforms \muon while reducing 40\% NS iterations across various scales and architectures. Now we verify that existing \muon variants cannot achieve these benefits at the same extent.

\subsubsection{Variants of Polar Method}
\begin{table}[t]
\centering
\small
\resizebox{\linewidth}{!}{%
\begin{tabular}{c|cc|cc}
\toprule
& \multicolumn{2}{c|}{\bf GPT-Small} & \multicolumn{2}{c}{\bf GPT-Base}\\
\midrule
\textit{NS Steps} & $N_s=3$ & $N_s=5$ & $N_s=3$ & $N_s=5$\\
\midrule
\cellcolor{blue!12}{\bf \framework} & \cellcolor{blue!12}{\bf 28.12} & \cellcolor{blue!12}{\bf 27.95} & \cellcolor{blue!12}{\bf 20.39} & \cellcolor{blue!12}{\bf 19.96} \\
\midrule
\makecell{\bf \muon\\\cite{jordan2024muon}} & \down{32.70}{4.58} & \down{29.51}{1.56}& \down{24.69}{4.30} & \down{21.47}{1.51} \\
\midrule
\makecell{\bf PolarExpress\\\cite{amsel2026polar}} & \down{30.01}{1.89} & \down{29.42}{1.47} & \down{22.74}{2.35} & \down{21.16}{1.20} \\
\midrule
\makecell{\bf Turbo-Muon\\\cite{boissin2025turbo}} & \down{29.70}{1.58} & \down{29.66}{1.71} & \down{23.46}{3.07} & \down{21.93}{1.97} \\
\midrule
\makecell{\bf NorMuon\\\cite{li2025normuon}} & \down{30.35}{2.23} & \down{28.40}{0.45} & \down{23.33}{2.94} & \down{21.27}{1.31} \\
\midrule
\makecell{\bf AdaMuon\\\cite{si2025adamuon}} & \down{31.20}{3.08} & \down{29.30}{1.35} & \down{26.07}{5.68} & \down{22.42}{2.46} \\
\bottomrule
\end{tabular}%
}
\caption{Validation perplexity ($\downarrow$) of \framework comparing against \muon variants on GPT-Small and Base.}
\label{tab:comparison-results}
\vspace{-5pt}
\end{table}


Methods that solely modify the polar approximation are closer to \framework in design principles, such as PolarExpress~\cite{amsel2026polar} and Turbo-Muon~\cite{boissin2025turbo}. We compare \framework with them on GPT-Small and GPT-Base. To ensure fairness, we also sweep hyper-parameters for baselines and report their best results in \cref{tab:comparison-results}. Full details are provided in \cref{apx:muon-variants-results}. By setting \framework as the baseline in \cref{tab:comparison-results}, we clearly observe that both PolarExpress and Turbo-Muon underperform \framework even with more NS iterations. They do suffer less degradation than plain \muon with reduced NS steps, but not comparable to \framework, and their improvement (if any) over \muon at $N_s=5$ is much less pronounced.

\subsubsection{Variants of Update Rules}
For other variants of \muon, we focus on ones that are closer to \framework in form, such as NorMuon~\cite{li2025normuon} and AdaMuon~\cite{si2025adamuon}. Despite of both using second moment in certain capacity, these approaches are fundamentally different from ours: they adapt the step size of each update direction similar to Adam, while \framework preserves the orthogonalized directions as in \muon.

Similarly, we compare \framework with NorMuon and AdaMuon on GPT-Small and GPT-Base with hyper-parameter sweeping, and report their best results in \cref{tab:comparison-results} (full results in \cref{apx:muon-variants-results}). Despite their improvement (if any) over \muon, both methods fail to reduce the necessary NS iterations, and they both underperform \framework in all settings, even with more NS steps than \framework.

\subsection{Training Speedup}
\begin{figure}[t]
    \centering
    \includegraphics[width=\linewidth]{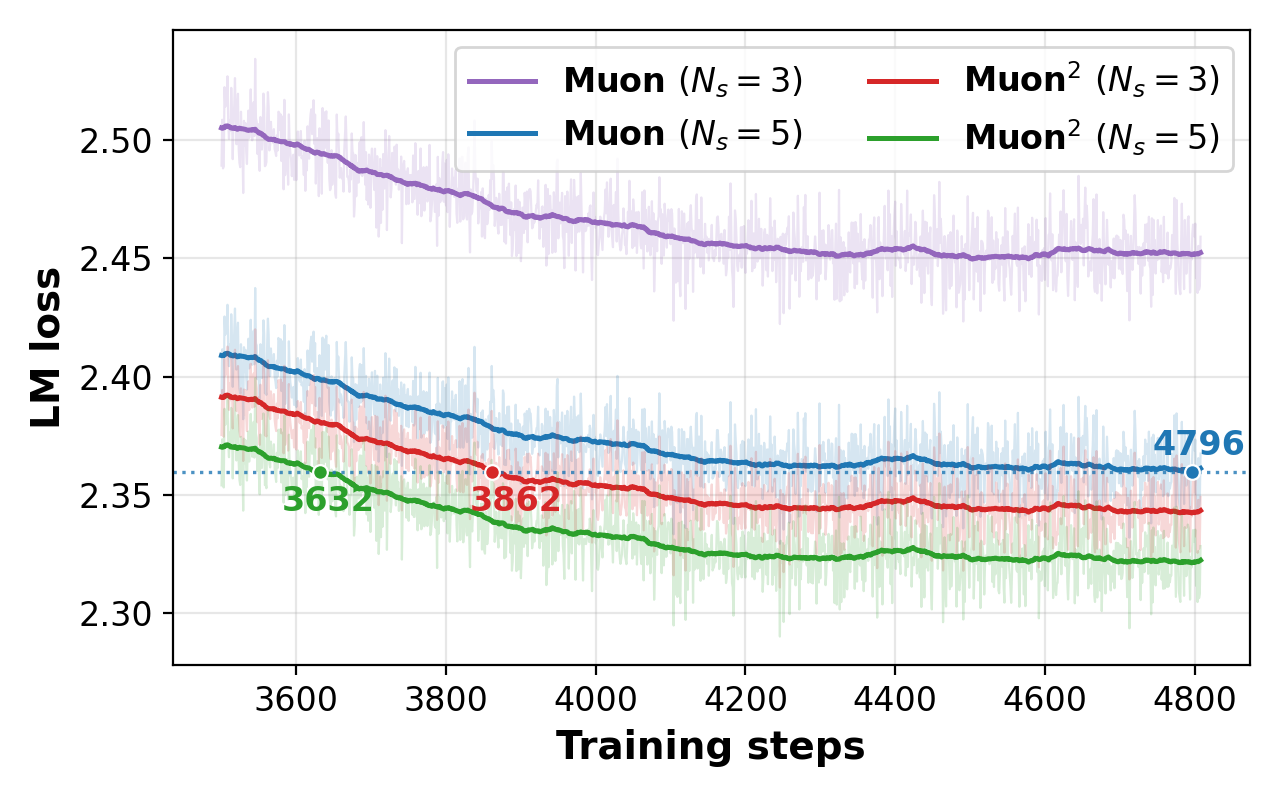}
    \caption{On LLaMA-1B, \framework achieves same loss with 25\% fewer training steps than \muon.}
    \label{fig:train-step-saving}
    \vspace{-10pt}
\end{figure}
We show in \cref{fig:train-step-saving} that on LLaMA-1B, \framework achieves the same training loss as \muon with up to 25\% fewer steps. Meanwhile, the second-moment scaling of \framework is computationally minimal, so that the per-training-step time is almost unchanged from \muon. As a result, the reduced steps directly translate to 25\% fewer training time (1042 vs. 796 GPU-hours) when achieving the same loss (2.36). For the same training steps, \framework with $N_s=3$ achieves a lower loss (2.34 vs 2.36) while reducing GPU-hours from 1044 to 1016. See \cref{apx:speedup}.

\subsection{Memory Efficiency of \meframework}
\begin{table}[t]
\centering
\footnotesize
\setlength{\tabcolsep}{5pt}
\resizebox{\linewidth}{!}{%
\begin{tabular}{c c c c c}
\toprule
\multirow{2}{*}{Model} & \multicolumn{4}{c}{Memory (GB)}\\
\cmidrule(lr){2-5}
& {\bf AdamW} & {\bf \muon} & {\bf \framework} & {\bf \meframework} \\
\midrule

LLaMA-1B & 20.39 & 18.56 & 20.36 & 18.57 \\
\midrule

LLaMA-7B & 59.31 & 47.29 & 60.66 & 47.53 \\
\bottomrule
\end{tabular}
}
\caption{Memory comparison across Adam, \muon, \framework and \meframework. }
\label{tab:memory}
\vspace{-10pt}
\end{table}

\cref{tab:memory} shows the memory usage across numerous approaches. \meframework uses practically same memory as \muon, verifying its superior memory efficiency while preserving most of \framework's performance gains. See details in \cref{apx:memory}.

\section{Conclusion}
We have presented \framework, a simple yet effective extension of \muon that applies adaptive second-moment preconditioning before orthogonalization, and shown that \framework achieves two desirable outcomes simultaneously: improved optimization behavior and reduced orthogonalization cost. We have analyzed the effect of \framework by first identifying the core challenge of the polar approximation lies in its ill-conditioned input, then showing how the proposed preconditioning fundamentally addresses this issue and results in significantly higher orthogonalization quality. As a result, \framework achieves practically sufficient orthogonalization with substantially fewer polar iterations. We have shown through comprehensive experiments that \framework(-F) consistently improves model performance while reducing polar iterations by 40\%. When achieving the same training loss, \framework saves up to 25\% GPU-hours over \muon, significantly lowers the massive pre-training cost, pushing forward the frontier of efficient and powerful optimizers for LLM pre-training.

\section{Limitation}
\framework is developed under academic budgets, following training settings in state-of-the-art literature~\cite{li2025normuon,si2025adamuon,zhang2026teon,ahn2025dion,amsel2026polar,boissin2025turbo}, where training models on 13B parameters already exceeds the scales adopted in most of these papers. As a result, we have not yet evaluated \framework at frontier-scale settings that are only viable under industrial budgets. In addition to \meframework, there may still exist other memory-efficient designs that better balance optimization quality and optimizer-state cost. We leave this as a future research.

\section{Ethical Considerations}
This work studies optimization methods for large-scale language model pre-training and does not introduce new datasets or involve the collection of personally identifiable information. All experiments are conducted on publicly available corpora~\cite{penedo2024fineweb,c4} that follow their respective licenses and terms of use. As our method focuses on improving training efficiency and optimization quality, we do not foresee ethical concerns beyond those already associated with large language model pre-training. In particular, models trained with \framework may still inherit biases, toxic content, hallucinations, or other harmful behaviors present in the underlying training data. Therefore, models trained using our optimizer should still undergo standard safety evaluation and responsible deployment practices.

\bibliography{acl_latex}

\clearpage
\appendix

\section{\meframework Algorithm}
Note that many approaches have been proposed to reduce the memory overhead introduced by the second-moment matrix~\cite{shazeer2018adafactor,adam-mini,zhao2024galore,chen2023symbolic,dettmers20218}, we leave it as future work for comprehensively studying and comparing among them. In this paper, we adopt Adafactor~\cite{shazeer2018adafactor} due to its simplicity and widely demonstrated effectiveness. The full algorithm of \meframework is provided in \cref{apx:muon2f-algo}.

\label{apx:muon2f-algo}
\begin{algorithm}[t]
\caption{The \meframework Optimizer}
\label{alg:muon2_adafactor}
\begin{algorithmic}[1]
\Require 2D weights $\mathbf{W}_t \in \mathbb{R}^{n\times m}$, objective $\mathcal{L}$, learning rate $\eta$, momentum coefficients $\beta_1,\beta_2$, Newton--Schulz steps $K$, numerical constant $\epsilon$
\Ensure Updated weights $\mathbf{W}_{t+1}$
\State $\mathbf{M}_0 \gets \mathbf{0}$, $\mathbf{r}_0 \gets \mathbf{0}\in\mathbb{R}^n$, $\mathbf{c}_0 \gets \mathbf{0}\in\mathbb{R}^m$
\For{$t \gets 1,2,\ldots$}
    \State $\mathbf{G}_t \gets \nabla_{\mathbf{W}_t}\mathcal{L}(\mathbf{W}_t)$
    \State $\mathbf{M}_t \gets \beta_1 \mathbf{M}_{t-1} + (1-\beta_1)\mathbf{G}_t$
    \State $\mathbf{R}_t \gets \mathbf{G}_t \odot \mathbf{G}_t$
    \State $\mathbf{r}_t \gets \beta_2 \mathbf{r}_{t-1} + (1-\beta_2)\,\mathrm{sum}_{\mathrm{col}}(\mathbf{R}_t)$
    \State $\mathbf{c}_t \gets \beta_2 \mathbf{c}_{t-1} + (1-\beta_2)\,\mathrm{sum}_{\mathrm{row}}(\mathbf{R}_t)$
    \State $\widehat{\mathbf{V}}_t \gets \dfrac{\mathbf{r}_t\,\mathbf{c}_t^\top}{\mathrm{sum}(\mathbf{r}_t)}$
    \State $\widetilde{\mathbf{M}}_t \gets \mathbf{M}_t \oslash (\sqrt{\widehat{\mathbf{V}}_t} + \epsilon \mathbf{1})$
    \State $\mathbf{O}_t \gets \mathrm{Newton\text{--}Schulz}(\widetilde{\mathbf{M}}_t, K)$
    \State $\mathbf{W}_{t+1} \gets \mathbf{W}_t - \eta\sqrt{m/n}\,\mathbf{O}_t$
\EndFor
\end{algorithmic}
\end{algorithm}

\section{Discussion on Cosine Similarity}
\label{apx:detail-cos-sim}

\begin{figure}[t]
    \centering
    \includegraphics[width=\linewidth]{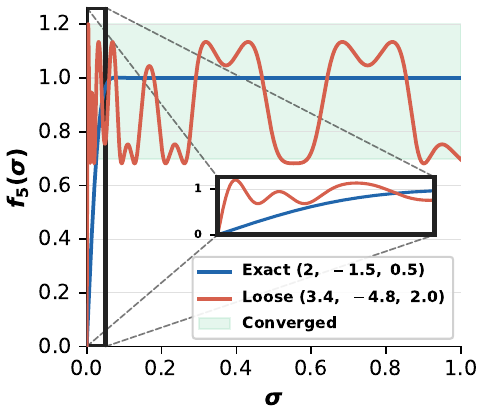}
    \caption{How the NS iteration maps singular values using different coefficients.}
    \label{fig:ns-mapping}
\end{figure}

To elaborate, cosine similarity [Eq.~\eqref{eq:cos-sim}] is robust against adversarial settings, such as in Eq.~\eqref{eq:scaled-polar} where a global scaling exists. Furthermore, it delivers an interpretable measurement that lies in $[0, 1]$ which reflects how much the polar approximate aligns with the ground truth direction, where 1 represents a perfect alignment and 0 represents an almost orthogonal direction. In contrast, the exact orthogonality error [Eq.~\eqref{eq:exact_orth}] gives a numerical value that is incident to the actual size of the matrix, and can only be used as a comparative metric for ranking purposes. Last but not least, cosine similarity presents a more practically meaningful measurement. To highlight how different messages these metrics convey, we consider a rather synthetic setting: we evaluate the singular values mapped by NS from a uniform grid of $[0,1]$, using the following two sets of coefficients:
\begin{itemize}[leftmargin=*]
    \item \emph{Loose target}: $(3.4445, -4.7750, 2.0315)$, which is adopted by~\cite{jordan2024muon} that roughly maps singular values from $[0, 1]$ to $[0.7, 1.3]$.
    \item \emph{Exact target}: $(2, -1.5, 0.5)$, an earlier attempt of~\cite{jordan2024muon} that vast majority of singular values are mapped to exact one except small values near zero.
\end{itemize}
\muon favors the \emph{loose target} as it rapidly reaches a practically sufficient orthogonalization with only 5 NS iterations (\cref{fig:ns-mapping}). However, 
Eq.~\eqref{eq:exact_orth} yields 0.03 for the \emph{exact target} and 0.31 for the \emph{loose target}, a 10x higher error for the latter. This reflects the fact that Eq.~\eqref{eq:exact_orth} measures the exact orthogonality, while in practice, the \emph{loose target} provides sufficient orthogonalization that is well aligned with the ground truth, suggested by the 0.98 cosine similarity given by Eq.~\eqref{eq:cos-sim}.

\section{Convergence Zones for PolarExpress}
\label{apx:zone-polarexpree}

\begin{figure}[]
    \centering
    \begin{subfigure}[t]{0.49\columnwidth}
        \centering
        \includegraphics[width=\linewidth]{fig/fig_spec_hist_step1.pdf}
        \caption{Keller}
        \label{fig:ns-in-sv-dist-early-pe}
    \end{subfigure}
    \hfill
    \begin{subfigure}[t]{0.49\columnwidth}
        \centering
        \includegraphics[width=\linewidth]{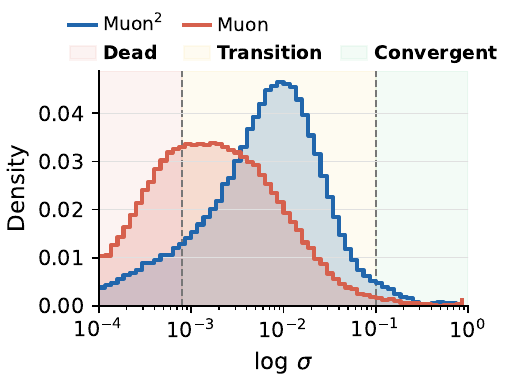}
        \caption{PolarExpress}
        \label{fig:ns-in-sv-dist-mid-pe}
    \end{subfigure}
    \caption{The comparison of convergence zones between Keller (\muon's~\cite{jordan2024muon} polar method) and PolarExpress~\cite{amsel2026polar}.}
    \label{fig:ns-in-spec-keller-vs-pe}
\end{figure}

As per discussion in \cref{sec:input-spectrum}, we divide the singular values of the NS input matrix from $[0, 1]$ into Dead, Transition, and Convergent zones. This partition is determined given particular choices of polar approximation method and the practical tolerance $\epsilon$, where the orthogonalization objective is mapping singular values to $[1-\epsilon, 1+\epsilon]$, as suggested by~\cite{jordan2024muon}. That said, the boundary of each convergence zone can change when different polar methods are used, such as PolarExpress~\cite{amsel2026polar}. When using PolarExpress, the boundary of dead zone reduces from 0.001 to 0.0008, and the boundary that separates transition and convergent zone reduces from 0.2 to 0.1. However, such changes are insufficient to solve the underlying challenge of the ill-conditioned input matrix. As demonstrate by \cref{fig:ns-in-spec-keller-vs-pe}, a substantial part of singular values for \muon with PolarExpress still fall in dead zone, and the advantages of \framework (i.e., significantly lower dead zone fraction, right-shifted distribution) persist.

\section{Proofs of \cref{prop:mp} and \cref{cor:signal}}
\label{apx:proofs}

Throughout, singular values are ordered decreasingly, $\sigma_1(A) \ge \cdots \ge \sigma_m(A) \ge 0$, and for an $m \times n$ matrix $A$ we write $F_{n,A}^{\sigma}(s) := \frac{1}{m}\#\{i : \sigma_i(A/\sqrt{n}) \le s\}$ for the empirical CDF of the singular values of $A/\sqrt{n}$. $G(s) := F^{\mathrm{MP}}(s^2)$ for $s \ge 0$ (and $0$ for $s<0$) denotes the singular-value form of the Marchenko--Pastur CDF $F^{\mathrm{MP}}$ with ratio $\lambda$. Since $\lambda < 1$, the Marchenko--Pastur density is bounded (its support $[(1-\sqrt{\lambda})^2, (1+\sqrt{\lambda})^2]$ is bounded away from zero), so $F^{\mathrm{MP}}$ and hence $G$ are continuous everywhere, with $G \equiv 0$ on $(-\infty, 1-\sqrt{\lambda}]$ and $G \equiv 1$ on $[1+\sqrt{\lambda}, \infty)$.

\subsection{Proof of \cref{prop:mp}}
\emph{Step 1 (spectrum under the $\sqrt{n}$ normalization).} By construction $\widetilde{M} = (\xi_{ij})$ has i.i.d.\ zero-mean unit-variance entries --- the scales cancel exactly, which is why every conclusion is independent of $(s_{ij})$. The almost-sure Marchenko--Pastur theorem \citep[Thm.~3.6]{bai2010spectral} applies to $S_n := \frac{1}{n}\widetilde{M}\widetilde{M}^\top$ and yields, on an event $\Omega_1$ of probability one, $F_m^{\mathrm{eig}}(x) \to F^{\mathrm{MP}}(x)$ for every $x$. Since $\sigma_i(\widetilde{M}/\sqrt{n}) = \sqrt{\lambda_i(S_n)}$ and squaring is strictly increasing on $[0,\infty)$, for $s \ge 0$ counting gives $F_{n,\widetilde{M}}^{\sigma}(s) = F_m^{\mathrm{eig}}(s^2) \to G(s)$; for $s < 0$, both $F_{n,\widetilde{M}}^{\sigma}(s)$ and $G(s)$ are zero. This proves the first claim.

\emph{Step 2 (Frobenius normalization).} Let $T_n := \|\widetilde{M}\|_F^2 = \sum_{ij} \xi_{ij}^2$, a sum of $N := mn$ i.i.d.\ terms with mean $1$ and finite variance $\mathbb{E}\xi^4 - 1$. Chebyshev's inequality gives $\mathbb{P}(|T_n/N - 1| \ge a) \le (\mathbb{E}\xi^4-1)/(a^2 mn)$, and since $m/n \to \lambda > 0$ implies $\sum_n \frac{1}{mn} < \infty$, the Borel--Cantelli lemma (applied for $a = 1/k$, $k \in \mathbb{N}$, then intersecting) yields $T_n/N \to 1$ almost surely; hence $R_n := \|\widetilde{M}\|_F/\sqrt{mn} \to 1$ on an event $\Omega_2$ of probability one. (If $\|\widetilde{M}\|_F = 0$, define the Frobenius-normalized matrix arbitrarily; since $R_n \to 1$ almost surely, this convention is eventually irrelevant.) The scaling rule $\sigma_i(cA) = c\,\sigma_i(A)$ applied twice gives the exact identity
\begin{equation*}
\sqrt{m}\,\sigma_i\big(\widetilde{M}/\|\widetilde{M}\|_F\big) = \sigma_i\big(\widetilde{M}/\sqrt{n}\big)\,/\,R_n .
\end{equation*}
On $\Omega_1 \cap \Omega_2$, fix $\delta \in (0,\tfrac12)$; for $n$ large, $1-\delta < R_n < 1+\delta$, so the empirical CDF of $\{\sigma_i/R_n\}$ at $s$ is sandwiched between $F_{n,\widetilde{M}}^{\sigma}(s(1-\delta))$ and $F_{n,\widetilde{M}}^{\sigma}(s(1+\delta))$; letting $n \to \infty$ and then $\delta \downarrow 0$, continuity of $G$ squeezes the limit to $G(s)$. Finally, for $t < 1-\sqrt{\lambda}$, the dead-zone fraction $\frac{1}{m}\#\{i : \sigma_i(\widetilde{M}/\|\widetilde{M}\|_F) \le t/\sqrt{m}\}$ equals this empirical CDF at $t$, which converges to $G(t) = 0$. \qed

\subsection{Proof of \cref{cor:signal}}
\emph{Case (i).} The rank inequality \citep[Thm.~A.44]{bai2010spectral} bounds the Kolmogorov distance between the eigenvalue ESDs of $\frac{1}{n}(D+\Xi)(D+\Xi)^\top$ and $\frac{1}{n}\Xi\Xi^\top$ by $\operatorname{rank}(D)/m$; since the eigenvalues are the squared singular values and $s \mapsto s^2$ is increasing on $[0,\infty)$, the same bound holds between the singular-value CDFs: $\sup_s |F_{n,D+\Xi}^{\sigma}(s) - F_{n,\Xi}^{\sigma}(s)| \le \operatorname{rank}(D)/m \to 0$, so the ESD of $(D+\Xi)/\sqrt{n}$ has the same limit $G$. For the normalization, the triangle inequality and Step 2 give $\limsup_n \|D+\Xi\|_F/\sqrt{mn} \le d + 1$ almost surely. Fix $t \ge 0$ with $t(1+d) < 1-\sqrt{\lambda}$ and $\varepsilon > 0$ small enough that $t(1+d+\varepsilon) < 1-\sqrt{\lambda}$; for $n$ large, the dead-zone fraction at threshold $t/\sqrt{m}$ is at most $F_{n,D+\Xi}^{\sigma}\big(t(1+d+\varepsilon)\big) \le F_{n,\Xi}^{\sigma}\big(t(1+d+\varepsilon)\big) + \operatorname{rank}(D)/m \to G\big(t(1+d+\varepsilon)\big) = 0$. \qed

\emph{Case (ii).} By Mirsky's inequality \citep{mirsky1960symmetric}, applied with both singular-value sequences in decreasing order (for rectangular matrices, zero-pad both to square ones; see also \citealp[Cor.~7.3.5]{horn2012matrix}), $\sum_i \big(\sigma_i((D+\Xi)/\sqrt{n}) - \sigma_i(\Xi/\sqrt{n})\big)^2 \le \|D\|_F^2/n = m\,\delta_n^2$ with $\delta_n^2 := \|D\|_F^2/(mn) \to 0$. Hence at most a fraction $\delta_n^2/\varepsilon^2$ of the indices satisfy $|\sigma_i^{D+\Xi} - \sigma_i^{\Xi}| \ge \varepsilon$, giving $F_{n,\Xi}^{\sigma}(s-\varepsilon) - \delta_n^2/\varepsilon^2 \le F_{n,D+\Xi}^{\sigma}(s) \le F_{n,\Xi}^{\sigma}(s+\varepsilon) + \delta_n^2/\varepsilon^2$; letting $n \to \infty$ and then $\varepsilon \downarrow 0$, continuity of $G$ gives $F_{n,D+\Xi}^{\sigma}(s) \to G(s)$ for every $s$. Moreover $\big|\|D+\Xi\|_F - \|\Xi\|_F\big| \le \|D\|_F = o(\sqrt{mn})$, so $\|D+\Xi\|_F/\sqrt{mn} \to 1$ almost surely, and the argument of Step 2 applies verbatim, yielding every conclusion of \cref{prop:mp}. \qed

\emph{Remark on the assumptions.} The i.i.d., zero-mean, unit-variance hypotheses enter through the Marchenko--Pastur theorem; the aspect ratio $m/n \to \lambda > 0$ enters both there and in Step 2, where it gives $\sum_n \frac{1}{mn} < \infty$; $\mathbb{E}\xi^4 < \infty$ enters only through the variance bound in Step 2 (without it, $R_n \to 1$ holds in probability and all distributional conclusions survive, with ``almost surely'' weakened accordingly); and $\lambda < 1$ buys the strictly positive lower edge $1-\sqrt{\lambda}$ --- at $\lambda = 1$ the limiting CDF is still continuous, but the edge sits at zero.

\emph{Remarks on the scope of the results.} (1) The guarantees are asymptotic and are stated at the threshold scale $t/\sqrt{m}$. To connect them with the empirical dead-zone boundary, consider the $512 \times 1344$ MLP projections of the LLaMA-60M measured in \cref{sec:input-spectrum}, where $\lambda \approx 0.38$: the limiting lower edge corresponds to $(1-\sqrt{\lambda})/\sqrt{m} \approx 1.7 \times 10^{-2}$ after the Frobenius normalization, an order of magnitude above the dead-zone boundary $10^{-3}$. (2) The claim in \cref{sec:theory} that an uneven scale profile can push a large fraction of singular values towards zero is verified by the following construction: if a fraction $\rho$ of rows of $M$ carries a negligible share of $\|M\|_F^2$, then zeroing these rows is a small Frobenius perturbation that reduces the rank to at most $(1-\rho)m$. Therefore, by Weyl's inequality for singular values, at least $\rho m$ singular values of the normalized raw matrix are correspondingly small. (3) \cref{prop:mp} divides by the exact scale $s_{ij}$, while \framework divides by its second-moment estimate. In the weak-signal regime of case (ii) in \cref{cor:signal}, the average signal-to-noise ratio $\frac{1}{mn}\sum_{ij}\mu_{ij}^2/s_{ij}^2$ vanishes, so $\mathbb{E}[M_{ij}^2] \approx s_{ij}^2$ for all but a vanishing fraction of entries and the estimate is consistent with the exact scale. Moreover, since the NS input is Frobenius-normalized, only the relative profile of the scales matters, and estimating it up to a common factor suffices. The additive $\epsilon$ in the estimate is orders of magnitude smaller than the entries of $\sqrt{\mathbf{V}_t}$ (see \cref{apx:sensitivity}) and therefore does not affect the profile. (4) When the scales factorize as $s_{ij} = r_i c_j$, which is the premise behind the factored second moment of \meframework, we have $D = \operatorname{diag}(1/r_i)\,(\mu_{ij})\,\operatorname{diag}(1/c_j)$, and multiplying by invertible diagonal matrices preserves rank. Therefore $\operatorname{rank}(D) = \operatorname{rank}((\mu_{ij}))$: the rank condition of case (i) in \cref{cor:signal} can be checked on the raw signal directly, while the bounded-strength condition remains required. (5) For square layers ($m = n$, $\lambda = 1$), the Marchenko--Pastur limit still applies, but its lower edge sits at zero and the positive margin of \cref{prop:mp} is lost. The same proof still shows that the fraction of singular values below $t/\sqrt{m}$ converges to the limiting CDF at $t$, which is at most $\frac{2}{\pi}t\,(1+o(1))$ under the singular-value law at $\lambda = 1$. Therefore, the preconditioned spectrum places no mass accumulation at zero, whereas an uneven scale profile can place a constant fraction of singular values near zero by the construction in (2).

\section{Detailed Evaluations}
\subsection{GPT Models}
\label{apx:gpt-sweep}
\begin{figure}[H]
    \centering
    \includegraphics[width=\linewidth]{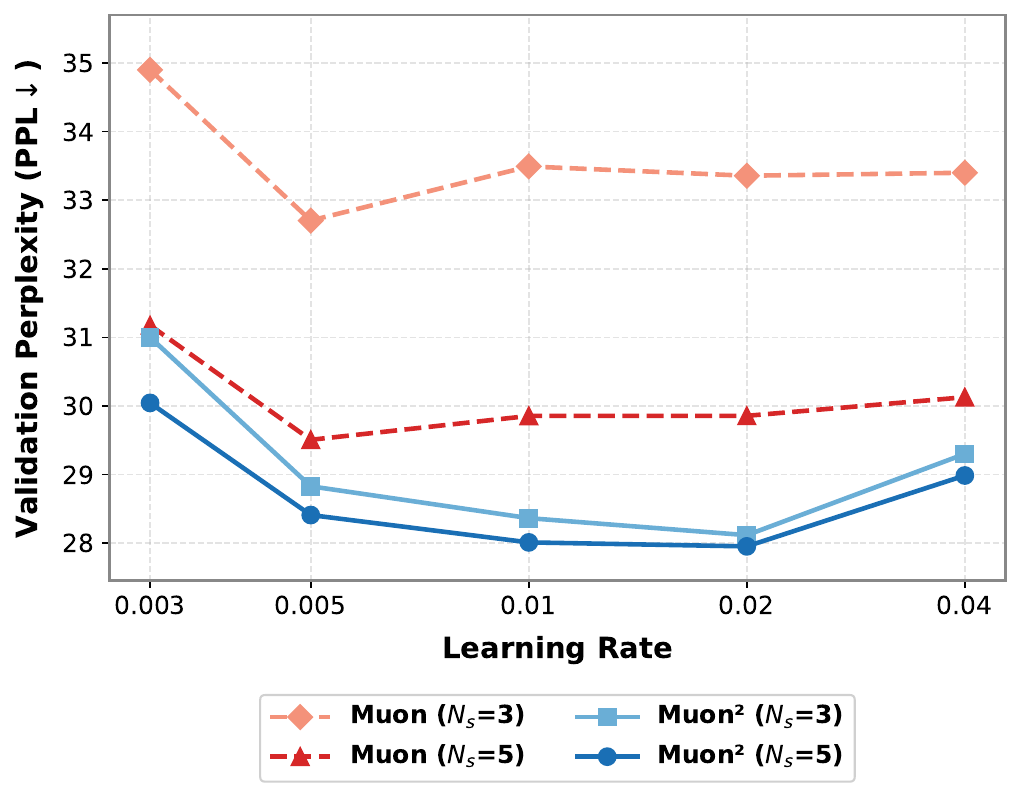}
    \caption{Learning Rates Sweep on {\bf GPT-Small}} 
    \label{fig:sweep_small_sweep}
\end{figure}

\begin{figure}[H]
    \centering
    \includegraphics[width=\linewidth]{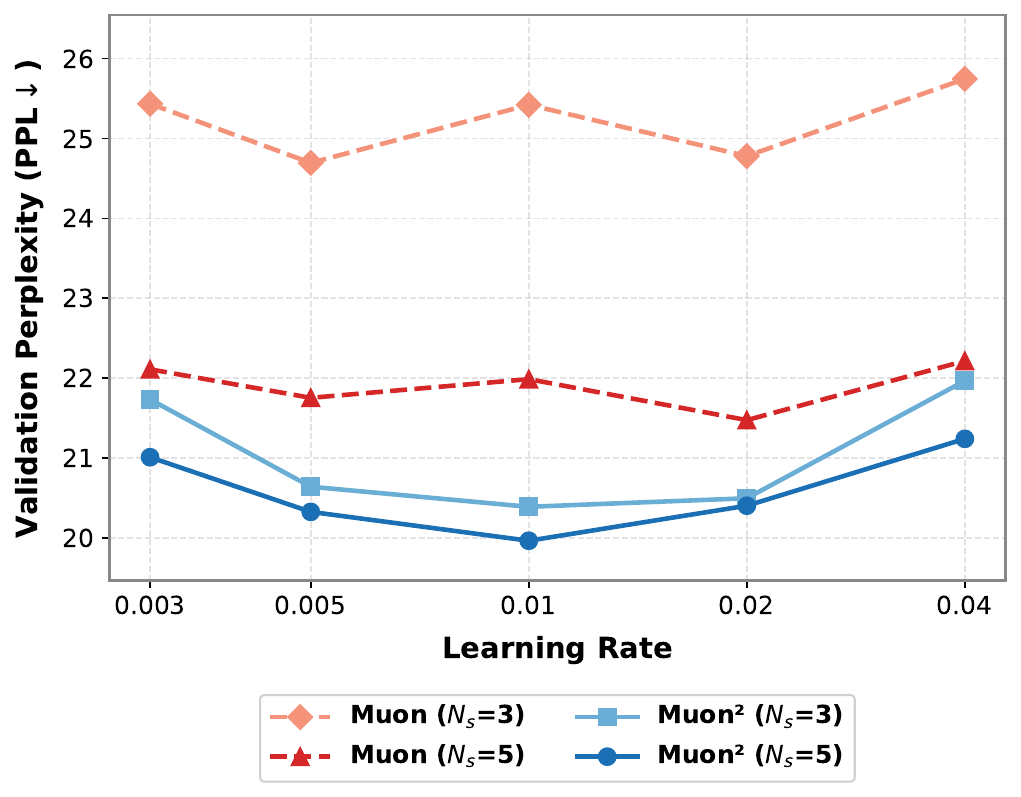}
    \caption{Learning Rates Sweep on {\bf GPT-Base}}
    \label{fig:sweep_base_sweep}
\end{figure}

Model configurations of each scale of the GPT models we considered are listed in \cref{tab:gpt_model_configs}. The learning rate sweeping results and their visualization are in \cref{tab:lr-sweep-small,fig:sweep_small_sweep} for GPT-Small, \cref{tab:lr-sweep-base,fig:sweep_base_sweep} for GPT-Base, and \cref{tab:lr-sweep-large,fig:gpt_large_sweep} for GPT-Large.

\begin{table}[H]
\centering
\resizebox{\linewidth}{!}{%
\begin{tabular}{lcccc}
\toprule
\textbf{Model} & $n_{\text{embd}}$ & $n_{\text{layer}}$ & $n_{\text{head}}$ & Param(M) \\
\midrule
GPT-Small & 768  & 12 & 12 & 124 \\
GPT-Base & 1024 & 24 & 16 & 362 \\
GPT-Large & 1280 & 36 & 20 & 774\\
\bottomrule
\end{tabular}
}
\caption{Architecture configurations of {\bf GPT} models.}
\label{tab:gpt_model_configs}
\end{table}

\begin{table}[H]
\centering
\setlength{\tabcolsep}{6pt}
\begin{tabular}{lcccc}
\toprule
\multirow{2}{*}{LR}
  & \multicolumn{2}{c}{Baseline}
  & \multicolumn{2}{c}{Muon$^2$ (ours)} \\
\cmidrule(lr){2-3}\cmidrule(lr){4-5}
  & $N_s{=}3$ & $N_s{=}5$ & $N_s{=}3$ & $N_s{=}5$ \\
\midrule
0.003 & 34.90 & 31.18 & {31.00} & {30.04} \\
0.005 & {\bf 32.70} & {\bf 29.51} & {28.83} & {28.41} \\
0.010 & 33.49 & 29.85 & {28.36} & {28.01} \\
0.020 & 33.36 & 29.86 & {\bf 28.12} & \textbf{27.95} \\
0.040 & 33.40 & 30.13 & {29.30} & {28.99} \\
\bottomrule
\end{tabular}
\caption{Learning rate sweep on \textbf{GPT-Small}. The best validation perplexity ($\downarrow$) of each method is bolded.}
\label{tab:lr-sweep-small}
\end{table}

\begin{table}[H]
\centering
\setlength{\tabcolsep}{6pt}
\begin{tabular}{lcccc}
\toprule
\multirow{2}{*}{LR}
  & \multicolumn{2}{c}{Baseline}
  & \multicolumn{2}{c}{Muon$^2$ (ours)} \\
\cmidrule(lr){2-3}\cmidrule(lr){4-5}
  & $N_s{=}3$ & $N_s{=}5$ & $N_s{=}3$ & $N_s{=}5$ \\
\midrule
0.003 & 25.43 & 22.11 & {21.73} & {21.01} \\
0.005 & {\bf 24.69} & 21.75 & {20.64} & {20.33} \\
0.010 & 25.42 & 21.99 & {\bf 20.39} & \textbf{19.96} \\
0.020 & 24.78 & {\bf 21.47} & {20.50} & {20.40} \\
0.040 & 25.75 & 22.22 & {21.97} & {21.24} \\
\bottomrule
\end{tabular}
\caption{Learning rate sweep on \textbf{GPT-Base}. The best validation perplexity ($\downarrow$) of each method is bolded.}
\label{tab:lr-sweep-base}
\end{table}

\begin{table}[H]
\centering
\setlength{\tabcolsep}{6pt}
\begin{tabular}{lcccc}
\toprule
\multirow{2}{*}{LR}
  & \multicolumn{2}{c}{Baseline}
  & \multicolumn{2}{c}{Muon$^2$ (ours)} \\
\cmidrule(lr){2-3}\cmidrule(lr){4-5}
  & $N_s{=}3$ & $N_s{=}5$ & $N_s{=}3$ & $N_s{=}5$ \\
\midrule
0.003 & 21.88 & 18.44 & 18.14 & {17.38} \\
0.005 & {\bf 21.13} & 18.32 & 17.28 & {16.82} \\
0.010 & 21.50 & 18.32 & {\bf 16.99} & {16.60} \\
0.020 & 21.57 & {\bf 17.57} & 17.28 & \textbf{16.52} \\
0.040 & 23.77 & 18.80 & 18.41 & {17.55} \\
\bottomrule
\end{tabular}
\caption{Learning rate sweep on \textbf{GPT-Large}. The best validation perplexity ($\downarrow$) of each method is bolded.}
\label{tab:lr-sweep-large}
\end{table}

\subsection{LLaMA Models}
\label{apx:llama-sweep}

\begin{figure}[H]
    \centering
    \includegraphics[width=\linewidth]{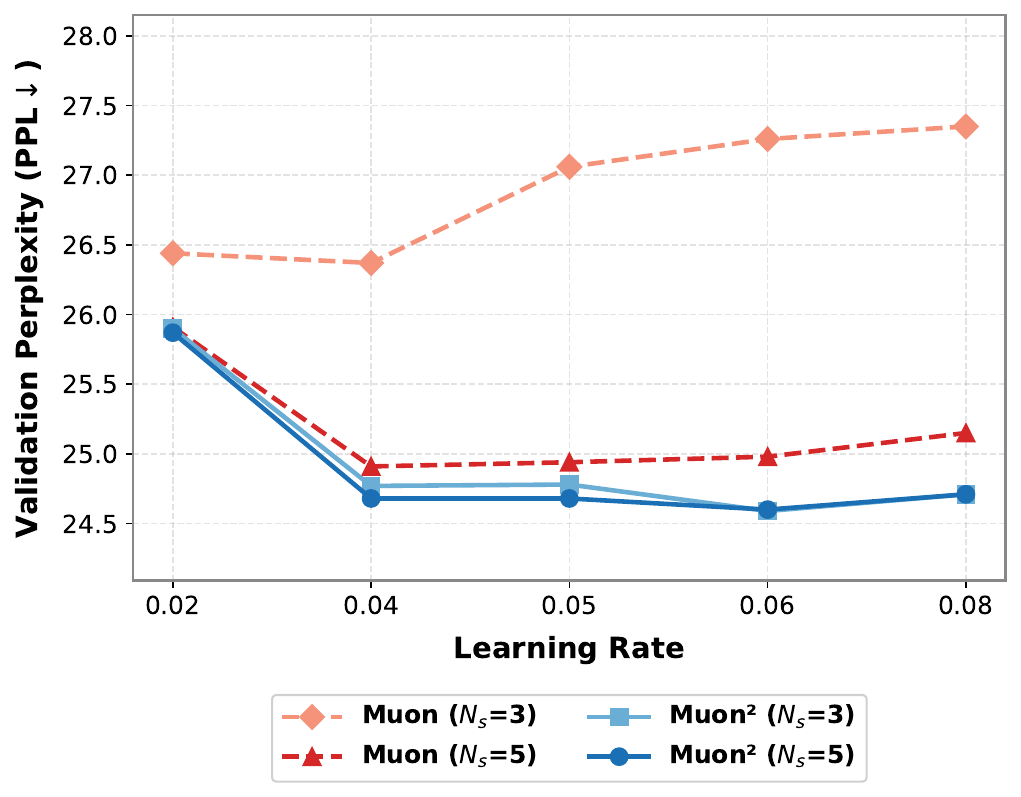}
    \caption{Learning Rates Sweep on {\bf LLaMA-60M}} 
    \label{fig:sweep_60m}
\end{figure}

\begin{figure}[H]
    \centering
    \includegraphics[width=\linewidth]{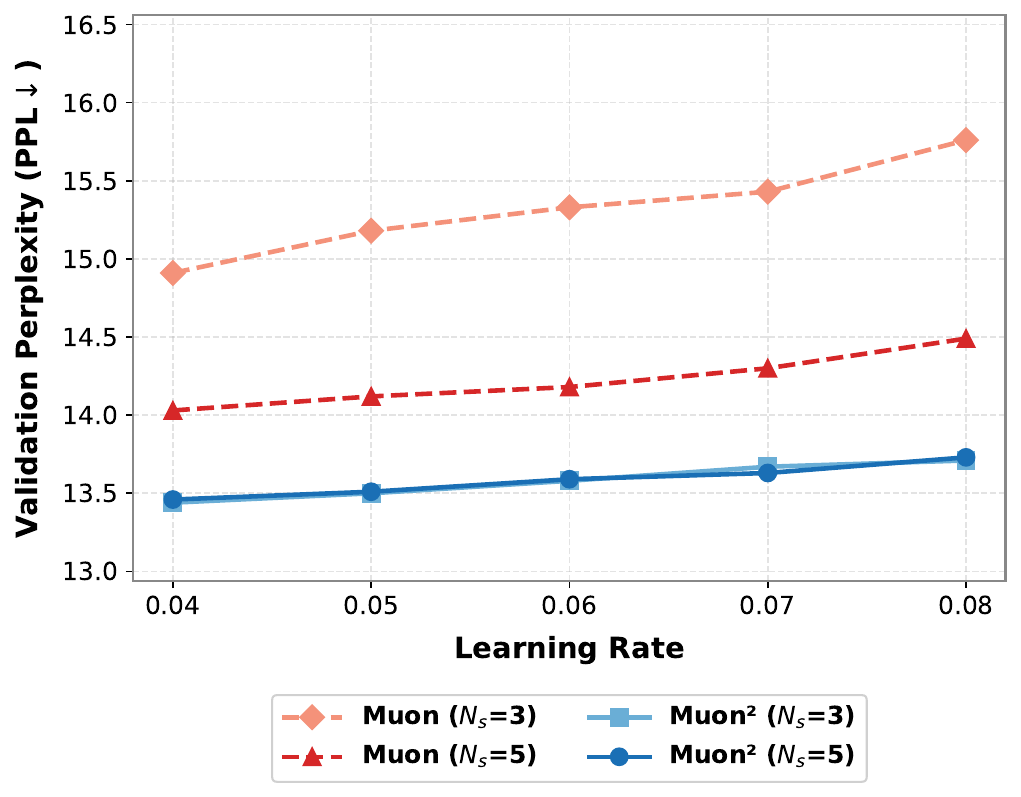}
    \caption{Learning Rates Sweep on {\bf LLaMA-350M}} 
    \label{fig:sweep_350m}
\end{figure}

\begin{figure}[H]
    \centering
    \includegraphics[width=\linewidth]{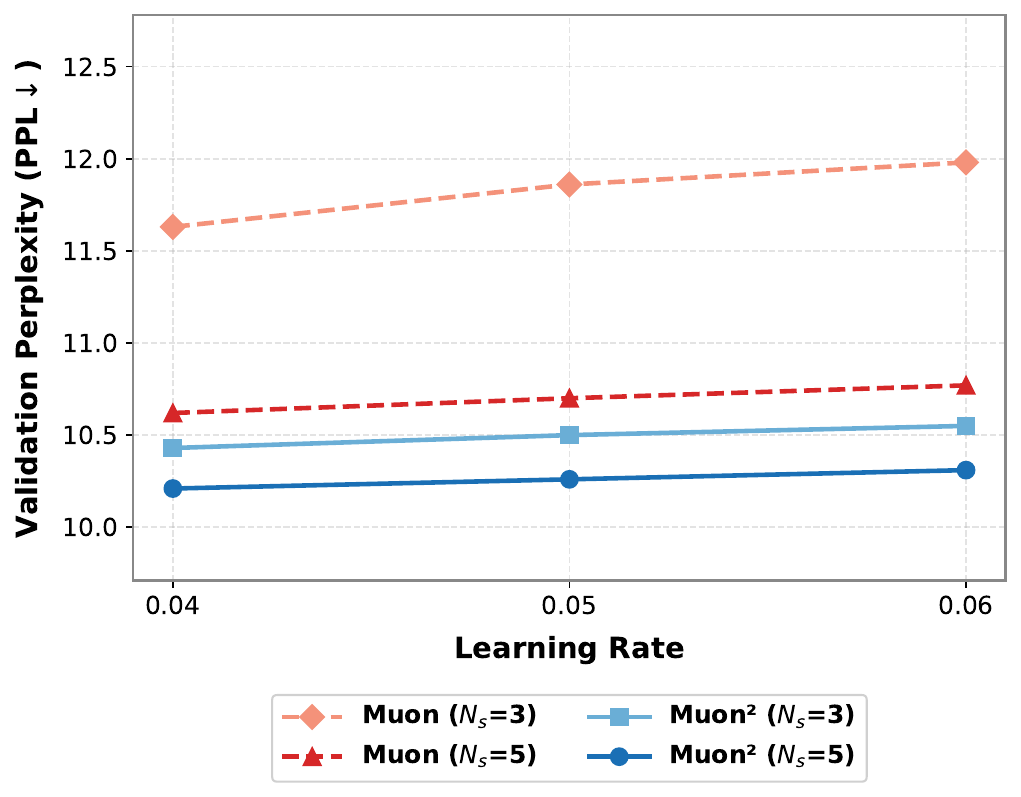}
    \caption{Learning Rates Sweep on {\bf LLaMA-1B}} 
    \label{fig:sweep_1b}
\end{figure}

Model configurations of each scale of the LLaMA models we considered are listed in \cref{tab:llama_model_configs}. The learning rate sweeping results and their visualization are in \cref{tab:lr-sweep-60m,fig:sweep_60m} for LLaMA-60M, \cref{tab:lr-sweep-350m,fig:sweep_350m} and for LLaMA-350M, and \cref{tab:lr-sweep-1b,fig:sweep_1b} for LLaMA-1B.

\begin{table}[H]
\centering
\resizebox{\linewidth}{!}{%
\begin{tabular}{lccccc}
\toprule
\textbf{Model} & $n_{\text{embd}}$ & $n_{\text{inter}}$ & $n_{\text{layer}}$ & $n_{\text{head}}$ & Param(M) \\
\midrule
LLaMA-60M & 512 & 1344 & 8 & 8 & 58 \\
LLaMA-350M & 1024 & 2736 & 24 & 16 & 368 \\
LLaMA-1B & 2048 & 5472 & 24 & 32 & 1280\\
LLaMA-13B & 5120 & 13827 & 40 & 40 & 13018 \\
\bottomrule
\end{tabular}
}
\caption{Architecture configurations of {\bf LLaMA} models.}
\label{tab:llama_model_configs}
\end{table}

\begin{table}[H]
\centering
\setlength{\tabcolsep}{6pt}
\begin{tabular}{lcccc}
\toprule
\multirow{2}{*}{LR}
  & \multicolumn{2}{c}{\muon}
  & \multicolumn{2}{c}{\framework (ours)} \\
\cmidrule(lr){2-3}\cmidrule(lr){4-5}
  & $N_s{=}3$ & $N_s{=}5$ & $N_s{=}3$ & $N_s{=}5$ \\
\midrule
0.02 & 26.44 & 25.91 & {25.90} & {25.87} \\
0.04 & {\bf 26.37} & {\bf 24.91} & {24.77} & {24.68} \\
0.05 & 27.06 & 24.94 & {24.78} & {24.68} \\
0.06 & 27.26 & 24.98 & {\bf 24.59} & \textbf{24.60} \\
0.08 & 27.35 & 25.15 & {24.71} & {24.71} \\
\bottomrule
\end{tabular}
\caption{Learning rate sweep on \textbf{LLaMA-60M}. The best validation perplexity ($\downarrow$) of each method is bolded.}
\label{tab:lr-sweep-60m}
\end{table}

\begin{table}[H]
\centering
\setlength{\tabcolsep}{6pt}
\begin{tabular}{lcccc}
\toprule
\multirow{2}{*}{LR}
  & \multicolumn{2}{c}{\muon}
  & \multicolumn{2}{c}{\framework (ours)} \\
\cmidrule(lr){2-3}\cmidrule(lr){4-5}
  & $N_s{=}3$ & $N_s{=}5$ & $N_s{=}3$ & $N_s{=}5$ \\
\midrule
0.04 & {\bf 14.91} & {\bf 14.03} & {\bf 13.44} & \textbf{13.46} \\
0.05 & 15.18 & 14.12 & {13.50} & {13.51} \\
0.06 & 15.33 & 14.18 & {13.58} & {13.59} \\
0.07 & 15.43 & 14.30 & {13.67} & {13.63} \\
0.08 & 15.76 & 14.49 & {13.71} & {13.73} \\
\bottomrule
\end{tabular}
\caption{Learning rate sweep on \textbf{LLaMA-350M}. The best validation perplexity ($\downarrow$) of each method is bolded.}
\label{tab:lr-sweep-350m}
\end{table}

\begin{table}[H]
\centering
\setlength{\tabcolsep}{6pt}
\begin{tabular}{lcccc}
\toprule
\multirow{2}{*}{LR}
  & \multicolumn{2}{c}{\muon}
  & \multicolumn{2}{c}{\framework (ours)} \\
\cmidrule(lr){2-3}\cmidrule(lr){4-5}
  & $N_s{=}3$ & $N_s{=}5$ & $N_s{=}3$ & $N_s{=}5$ \\
\midrule
0.04 & {\bf 11.63} & {\bf 10.62} & {\bf 10.43} & \textbf{10.21} \\
0.05 & 11.86 & 10.70 & {10.50} & {10.26} \\
0.06 & 11.98 & 10.77 & {10.55} & {10.31} \\
\bottomrule
\end{tabular}
\caption{Learning rate sweep on \textbf{LLaMA-1B}. The best validation perplexity ($\downarrow$) of each method is bolded.}
\label{tab:lr-sweep-1b}
\end{table}

\subsection{Mixture-of-Experts}
\label{apx:moe}
\begin{table}[H]
\centering
\resizebox{\linewidth}{!}{%
\begin{tabular}{lccccccc}
\toprule
\textbf{Model} & $n_{\text{embd}}$ & $n_{\text{inter}}$ & $n_{\text{layer}}$ & $n_{\text{head}}$ & $n_{\text{exp}}$ & $n_{\text{act}}$ & Param(M) \\
\midrule
MoE-7B-A1B & 2048 & 4096 & 24 & 32 & 16 & 2 & 6980 \\
\bottomrule
\end{tabular}
}
\caption{Architecture configurations of {\bf Mixture-of-Experts} model.}
\label{tab:moe_model_configs}
\end{table}

Model configuration of the Mixture-of-Experts model of \cref{sec:pretrain-moe} is in \cref{tab:moe_model_configs}.

\subsection{Downstream}
\label{apx:downstream}

We evaluate downstream performance on a standard suite of mostly zero-shot benchmarks covering commonsense reasoning, reading comprehension, and knowledge recall. Specifically, we report results on \textbf{ARC-Challenge} and \textbf{ARC-Easy}~\citep{clark2018think}, which test grade-school science questions at two difficulty levels; \textbf{OpenBookQA (OBQA)}~\citep{mihaylov2018can}, a multi-step reasoning benchmark requiring elementary science facts; \textbf{HellaSwag}~\citep{zellers2019hellaswag}, which evaluates commonsense sentence completion; \textbf{PIQA}~\citep{bisk2020piqa}, focused on physical commonsense reasoning; \textbf{WinoGrande}~\citep{sakaguchi2020winogrande}, a large-scale Winograd-style coreference resolution benchmark; \textbf{LAMBADA}~\citep{paperno2016lambada}, which measures long-range contextual word prediction; and \textbf{MMLU}~\citep{hendrycks2020measuring}, covering multitask language understanding across 57 subjects. We report accuracy for all tasks and the unweighted average (Avg) across the suite. See additional downstream evaluations on GPT-Large in \cref{tab:downstream-gpt2}.

\paragraph{Fine-tuned evaluations on LLaMA-1B.}
At this model scale, zero-shot accuracy on ARC-Challenge, WinoGrande, and MMLU remains close to the random-guessing baseline ($25\%$, $50\%$, and $25\%$, respectively), so zero-shot comparisons on these tasks carry little discriminative signal. For the three tasks marked with $^\dagger$ in \cref{tab:downstream-llama}, we therefore fine-tune each pretrained LLaMA-1B model on the corresponding training data before evaluation, applying the identical protocol to every optimizer. Fine-tuning minimizes the language-modeling loss on answer tokens only, using AdamW~\citep{loshchilov2017decoupled} with weight decay $0.01$, $\beta=(0.9, 0.95)$, a cosine schedule with $3\%$ warmup, gradient clipping at $1.0$, and bf16 mixed precision. For WinoGrande we use the official training set (40k examples; 2 epochs, batch size 32); for ARC-Challenge, the combined ARC-Easy and ARC-Challenge training splits (3.4k examples; 3 epochs, batch size 16); for MMLU, the auxiliary training set (77k examples truncated to 512 tokens; 1 epoch, batch size 32). For each model and task, the fine-tuning learning rate is swept over $\{0.5, 1, 2, 3, 5, 10\}\times 10^{-5}$ (up to $3\times 10^{-5}$ for MMLU), and we report the best configuration averaged over three random seeds. Evaluation follows the same harness as the zero-shot tasks~\citep[v0.4.12]{gao2021framework}: accuracy at 0-shot for ARC-Challenge and WinoGrande, and at 5-shot for MMLU following standard practice. \cref{tab:downstream-ft-lr} reports means and standard deviations across seeds for the sweep around the optimum: the sweep is first run with a single seed, and the two best learning rates per task are then repeated with three seeds. Comparing each model at its best learning rate, \framework consistently outperforms \muon at matched $N_s$ on all three tasks.

\paragraph{Fine-tuned evaluations on GPT-Large.}
We apply the same protocol to the GPT-Large models and mark the fine-tuned tasks with $^\dagger$ in \cref{tab:downstream-gpt2}, with two adaptations to the architecture: evaluation uses a context length of 1024, the maximum position of GPT-2, and MMLU fine-tuning enables activation checkpointing to keep batch size 32. For each task, the two learning rates that were optimal on LLaMA-1B are repeated with three seeds; \cref{tab:downstream-gpt-ft-lr} reports the sweep. The conclusions from LLaMA-1B carry over: \framework outperforms \muon at matched $N_s$ on ARC-Challenge and WinoGrande, and \framework at $N_s=3$ matches \muon at $N_s=5$ on ARC-Challenge while exceeding it on WinoGrande. The fine-tuning also resolves an artifact of the zero-shot table: the zero-shot WinoGrande column ranked \muon\ ($N_s=3$) highest (52.09), yet after fine-tuning the same model trails every other model by more than 4 points, confirming that rankings near the random baseline carry no signal. MMLU is the exception at this scale: after fine-tuning, all four models remain at the chance level ($25$--$27\%$), so we report the numbers for completeness but draw no comparison from MMLU on GPT-Large. We attribute this primarily to the model and data scale, as the larger LLaMA-1B with a 4096 context reaches only ${\sim}31\%$, and secondarily to the 1024 context, which truncates about $19\%$ of the 5-shot prompts.


\begin{table*}[t]
\centering
\small
\setlength{\tabcolsep}{5pt}
\renewcommand{\arraystretch}{1.15}
\resizebox{\linewidth}{!}{%

\begin{tabular}{lccccccc}
\toprule
& & \multicolumn{2}{c}{\textbf{ARC-c}} & \multicolumn{2}{c}{\textbf{WinoG.}} & \multicolumn{2}{c}{\textbf{MMLU}} \\
\cmidrule(lr){3-4} \cmidrule(lr){5-6} \cmidrule(lr){7-8}
& {\bf NS Steps} & $1{\times}10^{-5}$ & $2{\times}10^{-5}$ & $2{\times}10^{-5}$ & $3{\times}10^{-5}$ & $2{\times}10^{-5}$ & $3{\times}10^{-5}$ \\
\midrule
\multirow{2}{*}{\muon} & $N_s=3$ & $25.20_{\pm 0.05}$ & $\mathbf{27.08}_{\pm 0.26}$ & $56.70_{\pm 0.54}$ & $\mathbf{56.75}_{\pm 0.72}$ & $29.21_{\pm 0.86}$ & $\mathbf{29.75}_{\pm 0.32}$ \\
& $N_s=5$ & $29.27_{\pm 0.15}$ & $\mathbf{30.01}_{\pm 0.05}$ & $59.85_{\pm 0.32}$ & $\mathbf{59.91}_{\pm 0.83}$ & $30.24_{\pm 0.13}$ & $\mathbf{30.71}_{\pm 0.41}$ \\
\midrule
\multirow{2}{*}{\framework} & $N_s=3$ & $\mathbf{31.91}_{\pm 0.39}$ & $31.66_{\pm 0.39}$ & $\mathbf{59.88}_{\pm 0.93}$ & $59.30_{\pm 0.16}$ & $30.72_{\pm 0.51}$ & $\mathbf{30.79}_{\pm 0.82}$ \\
& $N_s=5$ & $32.65_{\pm 0.36}$ & $\mathbf{33.48}_{\pm 0.36}$ & $\mathbf{61.90}_{\pm 1.16}$ & $61.67_{\pm 1.13}$ & $\mathbf{31.31}_{\pm 0.78}$ & $30.60_{\pm 0.64}$ \\
\bottomrule
\end{tabular}
}
\caption{Fine-tuning learning-rate sweep on LLaMA-1B for the three fine-tuned tasks (mean and standard deviation over three seeds). For each model and task, the best configuration (bold) is the one reported in \cref{tab:downstream-llama}. Comparing each optimizer at its best learning rate, \framework outperforms \muon at matched $N_s$ on every task.}
\label{tab:downstream-ft-lr}
\end{table*}

\begin{table*}[t]
\centering
\small
\setlength{\tabcolsep}{6pt}
\renewcommand{\arraystretch}{1.15}
\resizebox{\linewidth}{!}{%
\begin{tabular}{lcccccccccc}
\toprule
& {\bf NS Steps} & \textbf{Avg} & \textbf{ARC-c}$^\dagger$ & \textbf{ARC-e} & \textbf{OBQA} & \textbf{HellaS.} & \textbf{PIQA} & \textbf{WinoG.}$^\dagger$ & \textbf{LAMBADA} & {\bf MMLU}$^\dagger$ \\

\midrule

\multirow{2}{*}{\muon}
& $N_s=3$ & 40.69 & 27.99 & 44.23 & 30.80 & 39.92 & 68.82 &
54.56 & 32.19 & 27.02 \\
& $N_s=5$ & 42.07 & 28.21 & 44.07 & 30.40 & 42.50 & 69.80 &
59.12 & 37.07 & 25.41 \\
\midrule

\multirow{2}{*}{\framework}
& \cellcolor{blue!6}$N_s=3$ & \cellcolor{blue!6}42.90 & \cellcolor{blue!6}28.30 & \cellcolor{blue!6}44.91 & \cellcolor{blue!6}29.40 & \cellcolor{blue!6}43.72 & \cellcolor{blue!6}69.70 &
\cellcolor{blue!6}\textbf{60.77} & \cellcolor{blue!6}40.03 & \cellcolor{blue!6}26.38 \\
& \cellcolor{blue!12}$N_s=5$ & \cellcolor{blue!12}\textbf{43.70} & \cellcolor{blue!12}\textbf{30.26} & \cellcolor{blue!12}\textbf{46.93} & \cellcolor{blue!12}\textbf{31.60} & \cellcolor{blue!12}\textbf{45.01} & \cellcolor{blue!12}\textbf{69.80} &
\cellcolor{blue!12}60.38 & \cellcolor{blue!12}\textbf{40.07} & \cellcolor{blue!12}25.52 \\
\bottomrule

\end{tabular}
}
\caption{Downstream evaluation on GPT-Large trained by \muon and \framework across different NS iterations. Tasks marked with $^\dagger$ are fine-tuned before evaluation under the protocol of \cref{apx:downstream}; the remaining tasks are zero-shot. MMLU is not bolded, as all four models remain at the chance level at this scale even after fine-tuning.}
\label{tab:downstream-gpt2}
\vspace{-10pt}
\end{table*}

\begin{table*}[t]
\centering
\small
\setlength{\tabcolsep}{5pt}
\renewcommand{\arraystretch}{1.15}
\resizebox{\linewidth}{!}{%

\begin{tabular}{lccccccc}
\toprule
& & \multicolumn{2}{c}{\textbf{ARC-c}} & \multicolumn{2}{c}{\textbf{WinoG.}} & \multicolumn{2}{c}{\textbf{MMLU}} \\
\cmidrule(lr){3-4} \cmidrule(lr){5-6} \cmidrule(lr){7-8}
& {\bf NS Steps} & $1{\times}10^{-5}$ & $2{\times}10^{-5}$ & $2{\times}10^{-5}$ & $3{\times}10^{-5}$ & $2{\times}10^{-5}$ & $3{\times}10^{-5}$ \\
\midrule
\multirow{2}{*}{\muon} & $N_s=3$ & $\mathbf{27.99}_{\pm 0.77}$ & $26.31_{\pm 0.68}$ & $\mathbf{54.56}_{\pm 0.75}$ & $53.85_{\pm 0.64}$ & $27.02_{\pm 0.55}$ & $26.62_{\pm 1.27}$ \\
& $N_s=5$ & $\mathbf{28.21}_{\pm 0.05}$ & $26.96_{\pm 0.31}$ & $58.93_{\pm 1.11}$ & $\mathbf{59.12}_{\pm 1.03}$ & $25.41_{\pm 0.24}$ & $25.22_{\pm 0.62}$ \\
\midrule
\multirow{2}{*}{\framework} & $N_s=3$ & $\mathbf{28.30}_{\pm 0.44}$ & $28.01_{\pm 1.02}$ & $\mathbf{60.77}_{\pm 0.60}$ & $60.43_{\pm 1.01}$ & $25.62_{\pm 0.11}$ & $26.38_{\pm 0.85}$ \\
& $N_s=5$ & $\mathbf{30.26}_{\pm 0.65}$ & $28.61_{\pm 0.43}$ & $60.27_{\pm 0.46}$ & $\mathbf{60.38}_{\pm 0.34}$ & $25.17_{\pm 0.46}$ & $25.52_{\pm 1.18}$ \\
\bottomrule
\end{tabular}
}
\caption{Fine-tuning learning-rate sweep on GPT-Large (mean and standard deviation over three seeds). For ARC-Challenge and WinoGrande, the best configuration per model (bold) is the one reported in \cref{tab:downstream-gpt2}. MMLU values are not bolded, as all of them remain at the chance level.}
\label{tab:downstream-gpt-ft-lr}
\end{table*}

\subsection{\muon Variants}
\label{apx:muon-variants-results}

We compare with \muon variants including PolarExpress~\cite{amsel2026polar}, Turbo-Muon~\cite{boissin2025turbo}, NorMuon~\cite{li2025normuon} and AdaMuon~\cite{si2025adamuon}. We focus on comparing them with \framework on GPT-Small and GPT-Base. We integrated their methods into our training framework using exactly what have been provided in their official repositories. For fairness, all methods are being sweeped on learning rate to make sure each method performs at their best capabilities, full sweeping results are in \cref{tab:lr-sweep-small-var,tab:lr-sweep-base-var,tab:lr-sweep-adamuon}. The reason we have AdaMuon separately in \cref{tab:lr-sweep-adamuon} is because it requires significantly smaller learning rates than other methods.

\begin{table}[H]
\centering
\setlength{\tabcolsep}{6pt}
\resizebox{\linewidth}{!}{%
\begin{tabular}{lcccccc}
\toprule
\multirow{2}{*}{LR}
  & \multicolumn{2}{c}{PolarExpress}
  & \multicolumn{2}{c}{Turbo-Muon}
  & \multicolumn{2}{c}{NorMuon}\\
\cmidrule(lr){2-3}\cmidrule(lr){4-5} \cmidrule(lr){6-7}
  & $N_s{=}3$ & $N_s{=}5$ & $N_s{=}3$ & $N_s{=}5$ & $N_s{=}3$ & $N_s{=}5$ \\
\midrule
0.003 & 31.51 & 30.45 & 31.36 & 30.31 & 32.50 & 29.90 \\
0.005 & {\bf 30.01} & {\bf 29.42} & 29.74 & {\bf 29.66} & 30.70 & {\bf 28.40} \\
0.010 & 31.32 & 29.73 & {\bf 29.70} & 29.79 & 31.43 & 28.72 \\
0.020 & 30.31 & 29.66 & 30.10 & 29.88 & 31.05 & 28.48 \\
0.040 & 31.25 & 30.03 & 30.07 & 30.82 & {\bf 30.35} & 29.09\\
\bottomrule
\end{tabular}
}
\caption{Learning rate sweep on \textbf{GPT-Small}. The best validation perplexity ($\downarrow$) of each method is bolded.}
\label{tab:lr-sweep-small-var}
\end{table}

\begin{table}[H]
\centering
\setlength{\tabcolsep}{6pt}
\resizebox{\linewidth}{!}{%
\begin{tabular}{lcccccc}
\toprule
\multirow{2}{*}{LR}
  & \multicolumn{2}{c}{PolarExpress}
  & \multicolumn{2}{c}{Turbo-Muon}
  & \multicolumn{2}{c}{NorMuon}\\
\cmidrule(lr){2-3}\cmidrule(lr){4-5} \cmidrule(lr){6-7}
  & $N_s{=}3$ & $N_s{=}5$ & $N_s{=}3$ & $N_s{=}5$ & $N_s{=}3$ & $N_s{=}5$ \\
\midrule
0.003 & 23.10 & 21.80 & 24.48 & 22.79 & 24.70 & 21.89 \\
0.005 & 22.75 & 21.39 & 23.95 & 22.51 & 23.90 & 21.54\\
0.010 & {23.38} & 21.44 & 24.14 & 22.42 & 24.18 & 21.42\\
0.020 & {\bf 22.74} & {\bf 21.16} & {\bf 23.46} & {\bf 21.93} & {\bf 23.33} & {\bf 21.27}\\
0.040 & 23.70 & 22.43 & 26.23 & 22.72 & 24.25 & 21.64 \\
\bottomrule
\end{tabular}
}
\caption{Learning rate sweep on \textbf{GPT-Base}. The best validation perplexity ($\downarrow$) of each method is bolded.}
\label{tab:lr-sweep-base-var}
\end{table}

\begin{table}[H]
\centering
\setlength{\tabcolsep}{6pt}
\begin{tabular}{lcccc}
\toprule
\multirow{2}{*}{LR}
  & \multicolumn{2}{c}{GPT-Small}
  & \multicolumn{2}{c}{GPT-Base} \\
\cmidrule(lr){2-3}\cmidrule(lr){4-5}
  & $N_s{=}3$ & $N_s{=}5$ & $N_s{=}3$ & $N_s{=}5$ \\
\midrule
0.001 & 36.63 & 31.49 & {\bf 26.07} & 22.73 \\
0.003 & {\bf 31.20} & {\bf 29.30} & 27.91 & 22.94 \\
0.005 & 33.45 & 30.48 & 26.08 & {\bf 22.42} \\
\bottomrule
\end{tabular}
\caption{Learning rate sweep of {\bf AdaMuon} on \textbf{GPT-Small} and {\bf GPT-Base}. The best validation perplexity ($\downarrow$) of each method is bolded.}
\label{tab:lr-sweep-adamuon}
\end{table}

\subsection{Sensitivity to $\beta_2$ and $\epsilon$}
\label{apx:sensitivity}

\framework introduces two hyperparameters over \muon: the second-moment decay $\beta_2$ and the numerical constant $\epsilon$. We assess sensitivity on LLaMA-60M under the paper's training setup, varying one at a time around the defaults ($\beta_2 = 0.95$, $\epsilon = 10^{-8}$); as a reference, we apply the identical protocol to \muon's own momentum coefficient. Results are in \cref{tab:sensitivity}.

\begin{table}[t]
\centering
\small
\setlength{\tabcolsep}{6pt}
\renewcommand{\arraystretch}{1.15}

\begin{tabular}{llc}
\toprule
& \textbf{Setting} & \textbf{Val.\ PPL} \\
\midrule
\multirow{5}{*}{\framework} & default ($\beta_2{=}0.95$, $\epsilon{=}10^{-8}$) & \textbf{24.59} \\
& $\beta_2 = 0.9$ & 24.72 \\
& $\beta_2 = 0.99$ & 24.69 \\
& $\epsilon = 10^{-6}$ & 24.74 \\
& $\epsilon = 10^{-10}$ & 24.56 \\
\midrule
\multirow{3}{*}{\muon} & momentum $= 0.95$ (default) & 24.98 \\
& momentum $= 0.9$ & 25.74 \\
& momentum $= 0.99$ & 26.63 \\
\bottomrule
\end{tabular}

\caption{Hyperparameter sensitivity on LLaMA-60M. Every \framework setting stays within $0.15$ PPL of the default and outperforms \muon, whereas \muon is substantially more sensitive to its own momentum coefficient.}
\label{tab:sensitivity}
\end{table}

\framework is robust to both hyperparameters: all settings fall within $0.15$ PPL of each other, and every setting outperforms base \muon (24.98). The robustness to $\epsilon$ is structural rather than tuned: $\epsilon$ enters only through the divisor $\sqrt{\mathbf{V}_t} + \epsilon\mathbf{1}$, so it can matter only when comparable to the entries of $\sqrt{\mathbf{V}_t}$; every conventional choice ($10^{-10}$ to $10^{-6}$) sits orders of magnitude below that scale. Unlike in Adam, $\epsilon$ also does not couple with the step size, because the NS iteration normalizes away the overall scale of its input. By contrast, \muon\ is markedly more sensitive to its momentum coefficient than \framework is to either $\beta_2$ or $\epsilon$.

\subsection{Ablation: Preconditioning without Orthogonalization}
\label{apx:no-ns}

To separate the improvement in NS convergence from a generic benefit of second-moment preconditioning, we ablate the NS iteration entirely: the preconditioned momentum $\widetilde{\mathbf{M}}_t$ is used directly as the update direction. At its best configuration on LLaMA-60M, this no-NS variant reaches a validation perplexity of 27.14 --- far behind both \framework (24.59) and base \muon (24.98). Second-moment preconditioning used directly as the update therefore does not deliver the gain; the improvement of \framework comes specifically from feeding the orthogonalization a better-conditioned input.

\subsection{Spectral Analysis on GPT-2 Small}
\label{apx:spectral-gpt2}

The spectral analyses of \cref{sec:input-spectrum} are conducted on a single model family. We repeat the full analysis on GPT-2 Small with identical definitions; \cref{tab:spectral-gpt2} summarizes the results. The conclusions carry over: \framework keeps roughly $4\times$ fewer singular values in the dead zone, about triples the effective rank of the NS input, and with 3 NS steps reaches the directional alignment that \muon\ needs 5 steps for (0.84 vs.\ 0.86). The gap is largest early in training: at step 50, $94.8\%$ of \muon's singular values fall in the dead zone (effective rank 2.1), versus $47.1\%$ for \framework (effective rank 8.5).

We additionally use this setting to examine whether update-scale fluctuations, which directional alignment deliberately ignores, could affect the effective learning rate: across 48 weight matrices and 38 logged steps, the coefficient of variation of the per-step update RMS is $0.06$ for \framework\ versus $0.20$ for \muon. This is expected: the update magnitude is set by the fixed post-NS rescaling shared by both methods, so any residual fluctuation stems from imperfect orthogonalization, of which \framework's better-conditioned input leaves less.

\begin{table}[t]
\centering
\small
\setlength{\tabcolsep}{6pt}
\renewcommand{\arraystretch}{1.15}

\begin{tabular}{lcc}
\toprule
\textbf{Metric} & \muon & \framework \\
\midrule
Dead-zone ratio & 0.26 & \textbf{0.06} \\
Transition-zone ratio & 0.73 & \textbf{0.93} \\
Effective rank & 48.9 & \textbf{142.5} \\
Directional alignment @ $N_s{=}1$ & 0.49 & \textbf{0.64} \\
Directional alignment @ $N_s{=}3$ & 0.69 & \textbf{0.84} \\
Directional alignment @ $N_s{=}5$ & 0.86 & \textbf{0.96} \\
\bottomrule
\end{tabular}

\caption{Spectral analysis of \cref{sec:input-spectrum} repeated on GPT-2 Small (identical definitions).}
\label{tab:spectral-gpt2}
\end{table}

\subsection{Training Speedup}
\label{apx:speedup}

\begin{figure}[H]
    \centering
    \includegraphics[width=\linewidth]{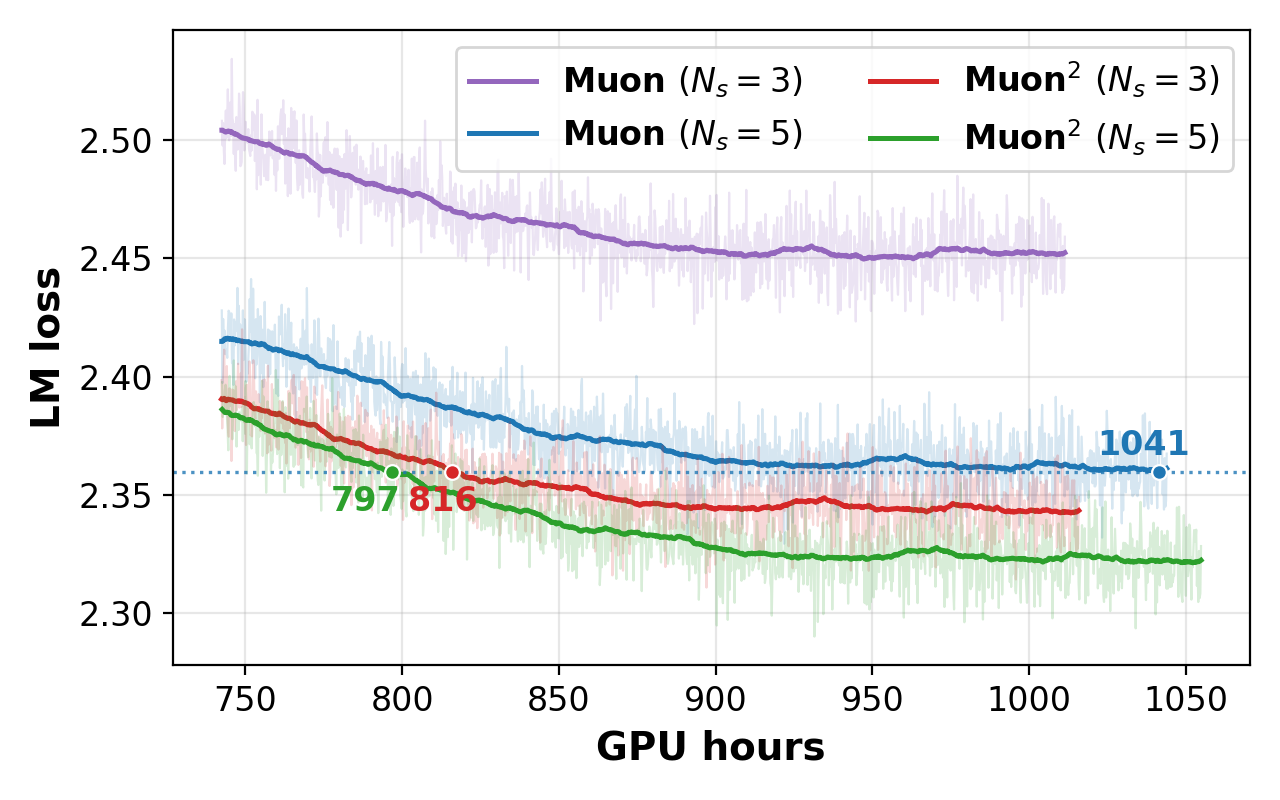}
    \caption{On LLaMA-1B, \framework achieves same loss with 1/4 fewer GPU-hours than \muon.} 
    \label{fig:walltime}
\end{figure}

Here we show the wall-time vs. training loss in \cref{fig:walltime}, and the per-training-step time in \cref{tab:step-time}.

\begin{table}[H]
\centering
\footnotesize
\setlength{\tabcolsep}{5pt}
\begin{tabular}{c c c}
\toprule
\textbf{Model} & {\bf NS Steps} & {\bf Step Time (ms)} \\
\midrule

\multirow{2}{*}{\muon}
& $N_s=3$ & 2958 \\
& $N_s=5$ & 3053 \\
\midrule

\multirow{2}{*}{\framework}
& $N_s=3$ & 2971 \\
& $N_s=5$ & 3085\\
\bottomrule
\end{tabular}
\caption{Per-training-step time for LLaMA-1B on a supercomputer with 64 nodes (4 GPUs each).}
\label{tab:step-time}
\end{table}

\subsection{Memory Efficiency of \meframework}
\label{apx:memory}
The memory profiling shown in \cref{tab:memory} is conducted on a single 94GB H100 GPU using training framework adopted from Nanotron~\cite{nanogpt} with bfloat 16 precision, sequence length 4096, and micro batch size 1.

\section{LLM Usage}
We used large language models solely as writing assistance tools for grammar and wording refinement. They were not involved in the development of ideas, experimental design, analysis, or interpretation of results. All technical content, claims, and conclusions were produced and verified by the authors.

\end{document}